\documentclass[9pt,twocolumn,twoside]{osajnl}
\usepackage{amsmath}
\usepackage{algorithm}
\usepackage{algpseudocode}
\usepackage{float}
\usepackage{placeins} 
\usepackage{lipsum}
\usepackage{stfloats}
\usepackage{graphicx} 
\usepackage{xcolor}
\usepackage{svg}

\journal{jocn} 

\setboolean{shortarticle}{false}

\title{From Data to Decision: A Multi-Stage Framework for Class Imbalance Mitigation in Optical Network Failure Analysis}

\author[1,*]{Yousuf Moiz Ali}
\author[1]{Jaroslaw E. Prilepsky}
\author[2]{Nicola Sambo}
\author[3,4]{Jo{\~a}o Pedro}
\author[5]{\\Mohammad M. Hosseini}
\author[5]{Antonio Napoli}
\author[1]{Sergei K. Turitsyn}
\author[1]{Pedro Freire}

\affil[1]{Aston Institute of Photonic Technologies, Aston University, B4 7ET, Birmingham, UK}
\affil[2]{Scuola Superiore Sant'Anna, 33 - 56127, Pisa, Italy}
\affil[3]{Nokia, Optical Networks, 2790-078 Carnaxide, Portugal}
\affil[4]{Instituto de Telecomunicações, Instituto Superior Técnico, 1049-001 Lisboa, Portugal}
\affil[5]{Nokia, Optical Networks, 81541 Munich, Germany}

\affil[*]{y.moizali@aston.ac.uk}

\begin{abstract}
Machine learning–based failure management in optical networks has gained significant attention in recent years, but severe class imbalance, where normal instances far outnumber failure cases, remains a considerable challenge. While pre- and in-processing techniques have been widely studied, post-processing methods are largely unexplored. We present a direct comparison of pre-, in-, and post-processing approaches for class imbalance mitigation in failure detection and identification using an experimental dataset. For failure detection, post-processing, particularly Threshold Adjustment, yields the highest F1 score improvement of up to 15.3\%, while Random Under-Sampling offers the fastest inference. In failure identification, Generative AI methods deliver the most significant performance gains up to 24.2\%, whereas post-processing has a limited impact in multi-class settings. When class overlap exists and latency is critical, over-sampling methods like Synthetic Minority Over-Sampling Technique (SMOTE) are most effective; without latency constraints, Meta-Learning excels. In low-overlap scenarios, Generative AI approaches provide the best performance with minimal inference time.

\end{abstract}

\setboolean{displaycopyright}{false} 

\begin{document}

\maketitle

\section{Introduction}
Optical networks are the foundation for a wide range of high-speed and latency-sensitive applications, such as video conferencing, real-time streaming, virtual reality, online gaming, etc.~\cite{dass2025heterogeneous}. The reliability and performance of these networks are therefore critical for both service providers and end users~\cite{ribeiro2025pca}. However, when data transmission failures occur, they can severely disrupt service delivery, leading to substantial financial losses and a degraded user experience~\cite{wang2022review}. As a result, effective and timely failure management in optical networks is essential to maintain service continuity and ensure the resilience of the modern communication infrastructure~\cite{musumeci2025failure}.

Historically, fiber-optic networks were designed to optimize cost, bandwidth, and transmission distance, while timely failure management was not the primary goal. 
Traditionally, failure management in optical networks has depended mainly on manual intervention and labor-intensive procedures~\cite{cruzes2024failure}. This approach is not only time-consuming but is also prone to delays in fault detection and service restoration, increasing the risk of prolonged service disruptions and reduced network reliability~\cite{wang2024alarmgpt, musumeci2025failure, cruzes2024failure}. Machine Learning (ML) has emerged as a path to automating and enhancing traditional manual failure management, addressing key limitations such as poor response times and vulnerability to human mistakes. ML-based approaches facilitate automatic response, proactive failure prediction, and faster fault detection by examining past network data and finding patterns associated with errors. Consequently, ML-based methods have recently attracted considerable interest in improving the robustness and effectiveness of optical network functions~\cite {musumeci2019tutorial}. 

At the same time, the introduction of ML has brought about its own set of challenges and problems. One of the most important and prevalent issues is the lack of good-quality datasets because network operators generally cannot share network data~\cite{musumeci2019tutorial}. The research community has made significant efforts to address this issue~\cite{akbari2025datasets, zhai2023open, santos2024experimental, bergk2021ml}. However, these studies are mainly based on datasets generated in simulated environments, which do not accurately reflect the characteristics of data collected from real-world optical networks. Another important issue is that even if a dataset is available, the distribution between normal and failure instances in the dataset is typically uneven, as normal (i.e., without failures) instances greatly outweigh the number of failure instances, since optical networks are designed to be robust~\cite{healy2025addressing}. This results in imbalanced training, which can lead to suboptimal performance; therefore, there is a clear need to find methods that efficiently tackle class imbalance and improve the performance of ML models. To address this problem, data-centric pre-processing techniques were considered in the context of failure detection and identification, including data augmentation and the generation of synthetic data using Generative AI (GenAI) methods, see Refs.~\cite{khan2022data, lun2023gan, khan2023data, sambo2023potential, kruse2024monitoring, healy2025addressing}. In these works, the authors have used GenAI models such as Generative Adversarial Networks (GANs) and Variational Autoencoders (VAEs) to generate synthetic samples. In~\cite{xing2023failure}, the authors have proposed the use of a time-series GAN (TimeGAN), a version of the conventional GAN tailored to time-series data. 

Model-centric in-processing approaches, which involve directly modifying the learning process of the ML algorithm, have also been explored in the literature~\cite{cichosz2021application, sun2023stacking, lun2023gan}. In~\cite{cichosz2021application}, the authors used one-class classification algorithms, such as one-class Support Vector Machine, one-class Naive Bayes classifier, and an Isolation Forest to focus only on the minority class. A Stacking Ensemble approach is proposed in~\cite{sun2023stacking} where the Long Short-Term Memory (LSTM) network and an XGBoost algorithm are used as the base models. These base models are trained on the original dataset, and preliminary predictions are generated, which then become the input of the final meta-learner; a meta-learner is typically a simple model such as Logistic Regression. In~\cite{lun2023gan}, the authors used a two-stage approach. In the first stage, they trained a GAN model on normal samples. During deployment, the GAN encoder and decoder are used to generate a latent representation of the original and decoded samples. The Euclidean distance between the two is used as a metric to detect failure. Similarly, for failure identification, they use a pre-trained Neural Network (NN) and a Gaussian Process Classification (GPC) algorithm. 

In some previous works, both pre-processing and in-processing techniques have been used. In~\cite{khan2024model}, the authors investigated Weighted Categorical Cross Entropy Loss, Focal Loss, Weighted Focal Loss, and Balanced Mini-Batch Training as in-processing techniques and Synthetic Minority Over-Sampling Technique (SMOTE), SMOTE-Tomek, and Conditional Tabular GAN (CTGAN) as pre-processing methods. These approaches were tested on an experimental dataset using an NN. In ~\cite{zhang2025shap}, a hybrid approach of Ensemble Learning (EL) was introduced at the model level and under-sampling at the data level was introduced in~\cite{zhang2025shap}. This was further validated by using a SHapley Additive exPlanations (SHAP) analysis to better explain the hybrid model and enable a more profound understanding of its working. Although these techniques improve the ML models, their effectiveness depends on the dataset used. If the dataset is not separable and has some degree of overlap between classes, GenAI models can struggle to generate good quality samples, which can lead to unsatisfactory results. To reduce the dependency on the quality of the dataset, post-processing or prediction-centric methods, which directly adjust the predictions from the ML model, can be efficient~\cite{siddique2023survey}.

This work presents a comprehensive comparison of pre-, in-, and post-processing class imbalance mitigation techniques on experimental datasets for failure detection and identification cases. In contrast to previous publications, which typically focus on a limited set of strategies to address class imbalance, our approach provides a systematic investigation of a wide range of techniques in the class imbalance mitigation paradigm. To the best of our knowledge, this study presents the most comprehensive evaluation of pre-processing, in-processing, and post-processing techniques in terms of the number and diversity of methods tested for class imbalance mitigation in the context of failure detection and identification. This work is also the first to investigate the effectiveness of post-processing methods for class imbalance mitigation in failure detection and identification. The novelty of our approach lies in its holistic scope, which offers a direct comparison across a wide range of approaches, enabling a clear understanding of which methods are effective under specific conditions and which may be less suitable, thereby offering practical guidance for future applications. In the pre-processing category, we tested traditional sampling methods such as SMOTE, Random Over-Sampling (ROS), and GenAI-based methods such as CTGANs and Conditional Variational Autoencoders (CVAEs). The in-processing strategies involve EL techniques such as Bagging and Boosting. For post-processing, the focus is on methods like Threshold Adjustment and Cost-Sensitive Thresholds. These methods will be explained in further detail in Section~\ref{sec:class_imb_miti}. Our results indicate that in a failure detection scenario, post-processing approaches such as Threshold Adjustment provide the most expressive improvement in terms of the F1-score. In contrast, the CTGAN offers the most improvement in F1-score in the failure identification case. 

The remainder of this paper is organized as follows. Section~\ref{sec:class_imb_miti} explains the different class imbalance mitigation techniques investigated in this study. Section~\ref{sec:exper_dataset_baseline} gives an outline of the experimental datasets used and establishes the baseline to which all techniques will be compared. Section~\ref{sec:results} presents and discusses the results of the detailed comparisons, and Section~\ref{sec:conclusion} concludes the article.

\section{Class Imbalance Mitigation Techniques}
\label{sec:class_imb_miti}

Class imbalance mitigation methods can be divided into three branches: pre-processing, in-processing, and post-processing~\cite{hort2024bias}. Pre-processing techniques consist of modifying the data before training. In-processing approaches update the model learning procedure, and post-processing techniques work with the predictions from the trained ML model to improve model performance~\cite{hort2024bias}. 

\subsection{Pre-processing}

Pre-processing strategies, also known as data-centric strategies, are techniques that modify the training set to counter class imbalance. These can be primarily divided into five categories: (i) over-sampling methods, (ii) under-sampling methods, (iii) a combination of over- and under-sampling, (iv) GenAI-based methods, and (v) heuristic-based label flipping. In the following sections, we will present these approaches in greater detail. 

\subsubsection{Over-sampling methods}
\begin{figure}[htbp]
\centering
\includegraphics[width=1.0\linewidth]{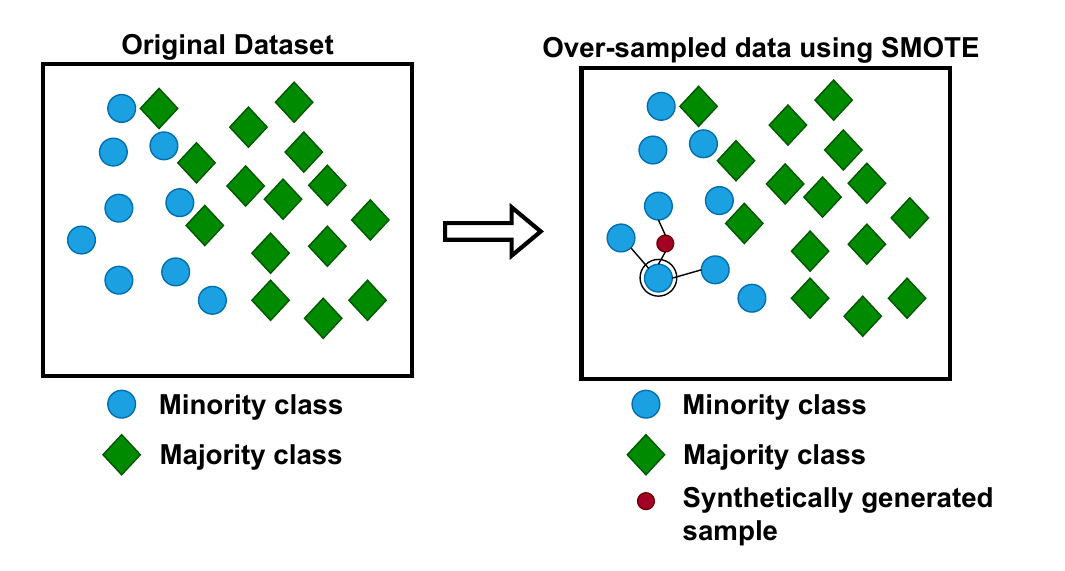}
\caption{Pictorial representation of SMOTE. (left) dataset before applying SMOTE, (right) dataset after applying SMOTE.}
\label{fig:smote}
\end{figure}

The over-sampling methods tested in this study include ROS, SMOTE, and the Adaptive Synthetic (ADASYN) over-sampling technique. ROS is a simple over-sampling technique in which the algorithm duplicates the minority class samples with replacement and does not change the class distribution~\cite{zhang2014rwo}. SMOTE was proposed in~\cite{chawla2002smote}. The idea is to randomly select a minority class sample from the original dataset and identify its \textit{k} nearest neighbors within the minority class with the help of the \textit{k}-Nearest Neighbors (kNN) algorithm~\cite{cunningham2021k}, where \textit{k} is a hyperparameter to tune. The final step involves generating synthetic samples by interpolating along the line segments connecting a minority class instance with one or more of its \textit{k} nearest neighbors. This can be visually understood with the help of Figure~\ref{fig:smote}. The plot on the left shows the imbalanced dataset before SMOTE. The plot on the right is the dataset after SMOTE, where a sample is generated synthetically by interpolating a minority class sample with one of its three nearest neighbors. ADASYN~\cite{he2008adasyn} also generates samples through interpolation, but the samples used to interpolate differ from those used by SMOTE. The ADASYN algorithm also uses the kNN classifier to find the \textit{k} nearest neighbors of the minority class samples, but it only generates synthetic samples for instances that have been incorrectly classified, thereby focusing on samples near the decision boundary~\cite{he2008adasyn}. It is important to note that ADASYN struggles to generate synthetic samples in cases where the minority and majority classes are well separated, since there are no areas where the minority class might be more challenging to learn. 

\subsubsection{Under-sampling methods}
This study focused on two under-sampling methods, Random Under-Sampling (RUS) and Cluster Centroids (CC). RUS is a way to balance the dataset by removing the samples of the majority classes at random until the desired balance between classes is reached~\cite{hasanin2018effects}. CC is another under-sampling method that makes a much more informed decision about which majority class samples to be removed~\cite{lin2017clustering}. The main idea behind CC is to apply a K-Means~\cite {sinaga2020unsupervised} clustering algorithm to establish the majority class clusters. The number of clusters is a hyperparameter to be tuned and can be selected on the basis of the distribution of the majority class samples. The next step is to find the cluster centroids of the identified clusters. The last step is to replace that entire cluster with the cluster centroid, thereby reducing the number of majority class samples.

\subsubsection{Combination of over-sampling and under-sampling}
\begin{figure}[htbp]
\centering
\includegraphics[width=1.0\linewidth]{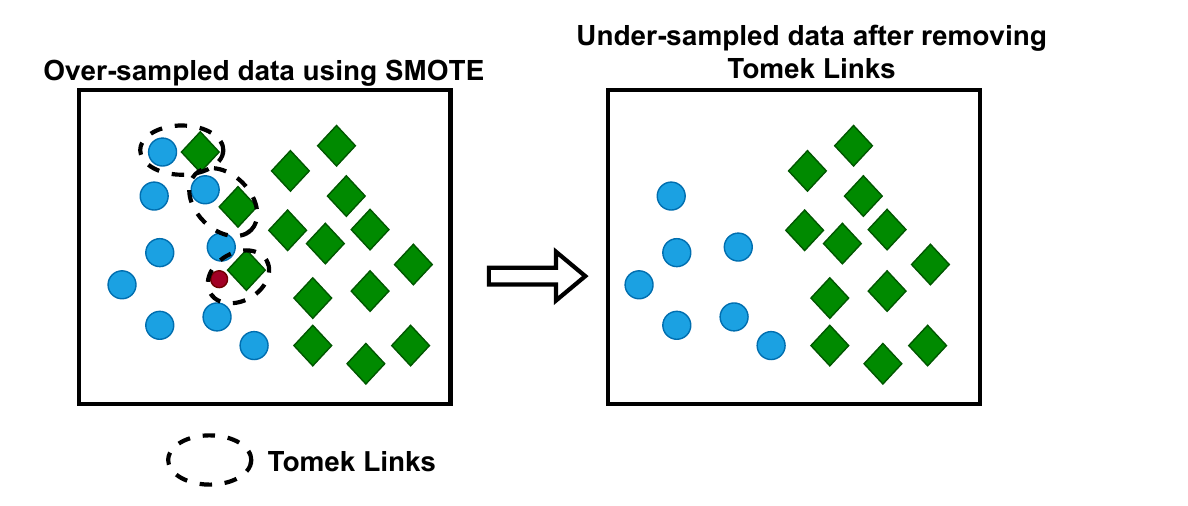}
\caption{Visual identification of the Tomek links on the plot on the left and their removal on the right.}
\label{fig:smote_tomek}
\end{figure}

The SMOTE-Tomek method provides a hybrid of over-sampling and under-sampling~\cite{batista2004study}. This technique is an extension of SMOTE, where the first step is to over-sample the data using SMOTE and then slightly under-sample the well-represented classes using Tomek links~\cite{tomek1976two}. Tomek links are formed when two samples from the opposite class are neighbors of each other. The samples that form Tomek links are removed to enhance the clarity and distinction of class boundaries. Figure~\ref{fig:smote_tomek} shows the SMOTE-Tomek technique, where the plot on the left-hand side shows the over-sampled data through SMOTE, as can be seen from a synthetically generated sample. Tomek links are then identified between opposite classes and removed, as seen in the right-hand plot. 

\subsubsection{Generative AI methods}
\begin{figure}[htbp]
\centering
\includegraphics[width=1.0\linewidth]{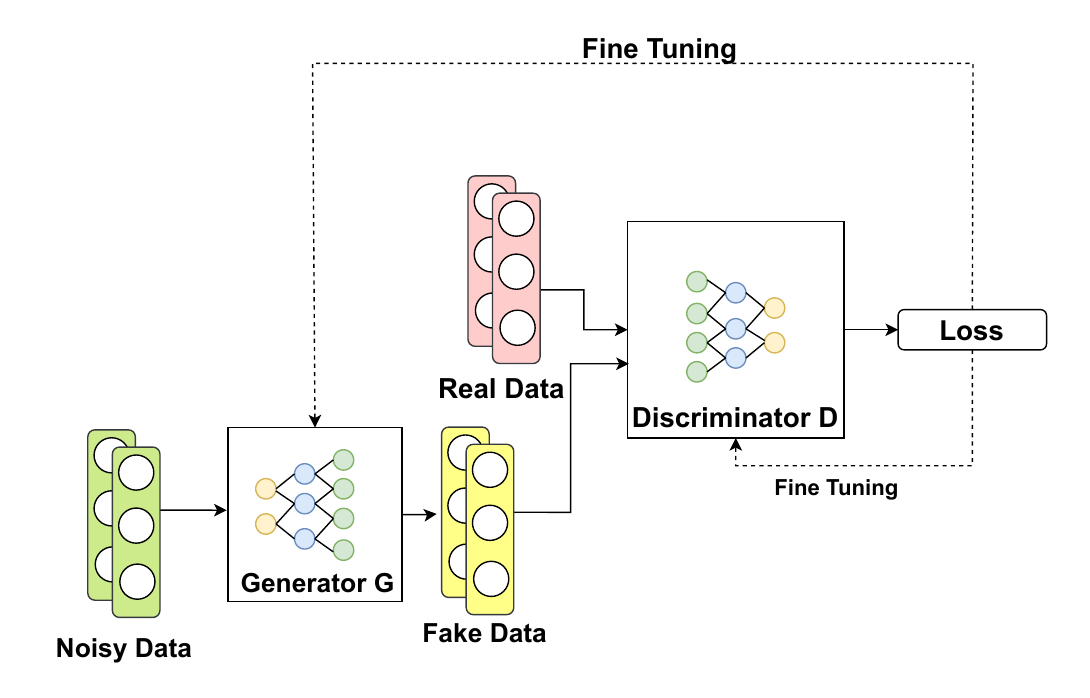}
\caption{A schematic for the GAN architecture (see Ref~\cite{saxena2021generative}).}
\label{fig:gan}
\end{figure}

Among GenAI methods, we tested CTGAN and CVAE. CTGAN is an extension of the conventional GAN algorithm. GAN was introduced in~\cite{goodfellow2014generative}. Figure~\ref{fig:gan} shows the architecture of the traditional GAN, which consists of two main parts, an NN-based generator and an NN-based discriminator. The idea is that the generator \textit{G} competes in an adversarial manner with the discriminator \textit{D}. The generator is fed with noisy data and outputs fake or synthetic data. The discriminator is then fed with real and fake samples to distinguish between both. The generator's goal is to generate good samples so that the discriminator can be fooled, and the discriminator's job is to distinguish fake data from real data correctly; thus, the term adversarial. 

However, conventional GAN suffers from a number of issues. One of the most prominent ones is the mode collapse issue, where the generator cannot generate data diverse enough, and the discriminator can easily separate real data from fake data~\cite{saxena2021generative}. Another very prominent issue is unstable training, since the generator and discriminator compete in an adversarial manner; there might be cases where the loss function fails to converge. GANs also struggle when working with tabular data, especially when it involves data of different types (i.e., numerical and categorical), and they ideally work best when working with a single datatype, such as images~\cite{saxena2021generative}. 

The authors in~\cite{ctgan} proposed the CTGAN approach, which is specifically designed for tabular data. In contrast to conventional GANs, CTGAN solves the problem of modeling mixed datatypes by utilizing the SoftMax activation for categorical variables and the $\tanh$ activation function for continuous ones at the output layer. The generation of conditional data is one of the key strengths of CTGAN, which allows synthetic samples to be generated based on specific target failure classes. Furthermore, the CTGAN also solves the common issue of the mode collapse problem by using the Wasserstein~\cite{arjovsky2017wasserstein} loss function. Considering these advantages, CTGAN is a more suitable choice for generating synthetic samples in optical network datasets, which are predominantly tabular in nature.

\begin{figure}[htbp]
\centering
\includegraphics[width=1.0\linewidth]{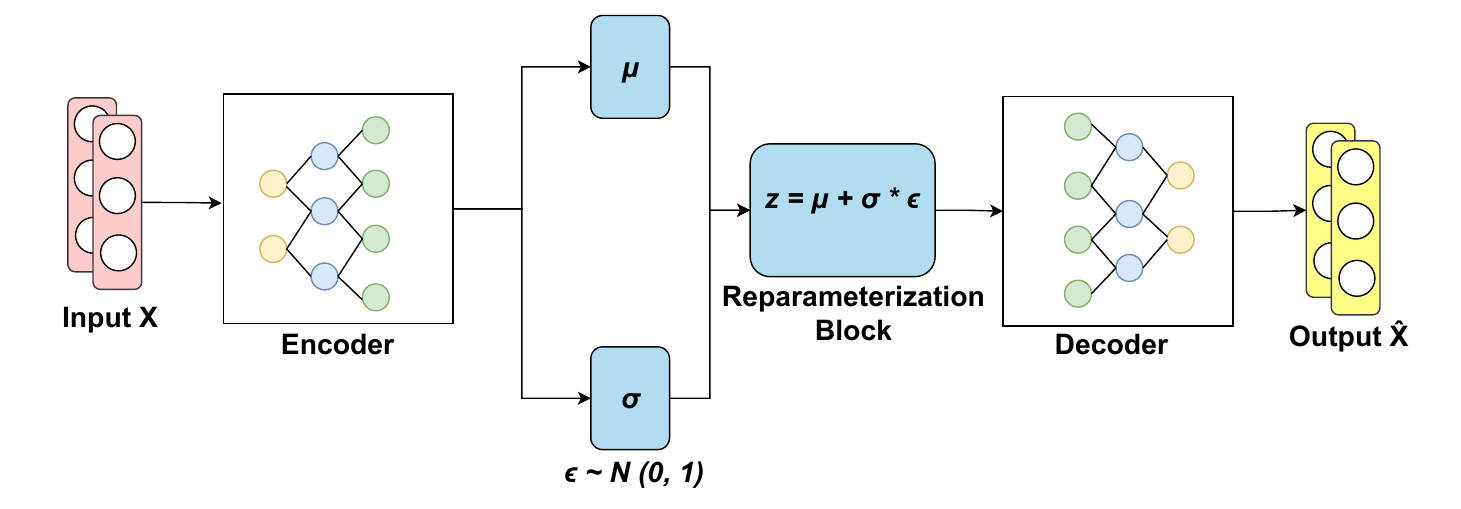}
\caption{A schematic for the VAE architecture (see Ref~\cite{doersch2016tutorial}).}
\label{fig:vae}
\end{figure}

Similar to the case of the CTGAN, CVAE is an extension of the VAE for tabular data. Figure~\ref{fig:vae} provides an overview of the VAE architecture. Like GAN, it contains an NN-based encoder and an NN-based decoder~\cite{kingma2013auto}. The main idea behind the VAE is to learn a probabilistic latent representation of the input data. The job of the encoder is to compress the input into a low-dimensional latent space, which is characterized by the mean $\mu$ and standard deviation $\sigma$. The model then samples a latent variable \textit{z} from the distribution rather than encoding a single point with the help of the \textit{Reparameterization Block}, as seen in Figure~\ref{fig:vae}. This block is an essential part of the VAE, as it enables backpropagation to be carried out through the NN. The function of the decoder is to reconstruct the original input data from this low-dimensional representation. Since the distribution of the latent space is known beforehand, the encoders and decoders are said to be probabilistic in nature, making them different from traditional autoencoders. By learning a structured latent space aligned with a known prior (typically a standard normal distribution), the VAE enables both reconstruction and generation of new, coherent samples. In this work, we use CVAE, which is a version of VAE suitable for the generation of conditional data~\cite{ctgan}. Similar to the CTGAN case, CVAE enables us to generate data for specific failure classes, which is ideal in the context of failure detection and identification. 
\subsubsection{Heuristics-based label flipping}
In this category, we tested Massaging~\cite{hort2024bias}, Perturbation~\cite{hort2024bias}, and Cluster-based Massaging. Massaging is a simple label-flipping technique, in which we flip the labels of those majority class instances with a high probability of belonging to the minority class or the failure class. These samples are usually found close to the decision boundary of the model. The idea behind Perturbation can be explained by the following formula:
\begin{equation}
f(x) = 
\begin{cases}
1, & \text{if } x = 0 \text{ and } \text{rand}() < p_{\text{majority\_flip}}, \\
0, & \text{if } x = 1 \text{ and } \text{rand}() < p_{\text{minority\_flip}}, \\
x, & \text{otherwise}.
\end{cases}
\label{eq:perturbation}
\end{equation}
We generate a random number for every sample \textit{x} in our dataset. Assuming $0$ is the label for the majority class and $1$ is the label for the minority class, if $x = 0$ and the generated random number is less than $p_{\text{majority\_flip}}$, we flip this label to $1$. Similarly, if $x = 1$ and the random number generated is less than $p_{\text{minority\_flip}}$, we will flip this label to $0$. Typically, $p_{\text{majority\_flip}}$ is set to a higher value than $p_{\text{minority\_flip}}$ to encourage more labels of the majority class to flip. This is a hyperparameter that can be tuned. Note that this explanation is for the binary case (i.e., failure detection), but it can be easily extended for the multi-class case (i.e., failure identification). In the multi-class case, we have to focus on labels with a significant amount of overlap and consider only flipping the labels of those classes because flipping the labels of well-separated classes would only introduce noise in the dataset.

Lastly, Cluster-based Massaging is an extension of the original Massaging method. The first step is to divide the majority class into clusters using the K-Means algorithm~\cite{sinaga2020unsupervised}. The number of clusters is a hyperparameter and can be tuned based on the distribution of the majority class. The next step is to find the centroid of the minority class that is under consideration. Then we calculate the Euclidean distance of the centroid to each of the majority class clusters. The cluster closest to the minority class is selected, and a small portion (typically 20\%) of the samples in the cluster is flipped to the minority class label. 

Practically, over-sampling approaches, including GenAI-based data augmentation, are widely used in classification, as they preserve information and boost minority recall. However, one must guard against overfitting or noise creation. Under-sampling is useful in scenarios when data are plentiful or the majority class contributes to significant noise in the dataset to the point that under-sampling actually improves the data distribution, as it accelerates training, but risks discarding useful patterns. A hybrid methodology, such as SMOTE-Tomek, frequently yields optimal results in structured data. Heuristics-based methods work best in binary scenarios where it is much easier to identify which labels to flip. In short, use over-sampling when minority data are scarce and the feature space is relatively low-dimensional, and use under-sampling when majority data dominates noise or computation time is of utmost importance.

\subsection{In-processing}
In-processing techniques, also known as model-centric approaches, adapt the model or loss function to mitigate class imbalance. They do not alter the training dataset, yet they concentrate on the minority samples during training. We tested five approaches in this domain: Weighted Learning, Ensemble Learning, Balanced Random Forests, Balanced Epoch Training, and Meta-Learning. The following paragraphs explain these methods in more detail. 

\subsubsection{Weighted Learning}
Weighted learning~\cite{bakirarar2023class} is a common class imbalance mitigation technique because it is relatively easy to implement. The idea is to assign different weights to the different classes during the training time. The model prioritizes minority classes by assigning them higher weights, whereas majority classes have reduced weights, enhancing focus during training. The weight of the classes is calculated by the formula in Eq.~(\ref{eq:cw})
\begin{equation}
CW_i = \frac{n_{\text{samples}}}{n_{\text{classes}} \cdot {n_{\text{samples}}}_i}.
\label{eq:cw}
\end{equation}
To calculate the weight of class \textit{i}, $CW_i$, divide the total number of samples, ${n_{\text{samples}}}$, into the product of the total number of classes, ${n_{\text{classes}}}$, and the number of samples in class \textit{i}, ${n_{\text{samples}}}_i$. These weights are directly incorporated into the loss function of the algorithm under consideration. 

We also implemented the Weighted Focal Loss (WFL) technique for using NNs in failure identification. The WFL is an extension of the original Focal Loss (FL) introduced in~\cite{lin2017focal}. The standard cross-entropy loss used in NNs treats all samples equally, regardless of their difficulty. As a result, easily classified samples can disproportionately influence the loss and dominate the gradient, potentially hindering the model's ability to learn from more complex and informative examples. The FL overcomes this problem by allowing the model to focus less on easily classified samples and more on harder-to-classify samples. The FL loss is shown in the following formula:
\begin{equation}
\text{FL}(y_{\text{true}}, y_{\text{pred}}) = - y_{\text{true}} \cdot (1 - y_{\text{pred}})^\gamma \cdot \log(y_{\text{pred}}).
\label{eq:fl}
\end{equation}
The term $(1 - y_{\text{pred}})^\gamma$ is the modulating factor, and $\gamma$ is the tunable focusing parameter. The WFL loss is shown below:
\begin{equation}
\text{WFL}(y_{\text{true}}, y_{\text{pred}}) = - \alpha \cdot y_{\text{true}} \cdot (1 - y_{\text{pred}})^\gamma \cdot \log(y_{\text{pred}}) ,
\label{eq:wfl}
\end{equation}
where the only addition is the tunable parameter $\alpha$. This addition is particularly beneficial in imbalanced classification tasks, as it allows the model to assign higher importance to minority classes through appropriately tuned $\alpha$ values.

\subsubsection{Ensemble Learning (EL)}
\begin{figure}[htbp]
\centering
\includegraphics[width=1.0\linewidth]{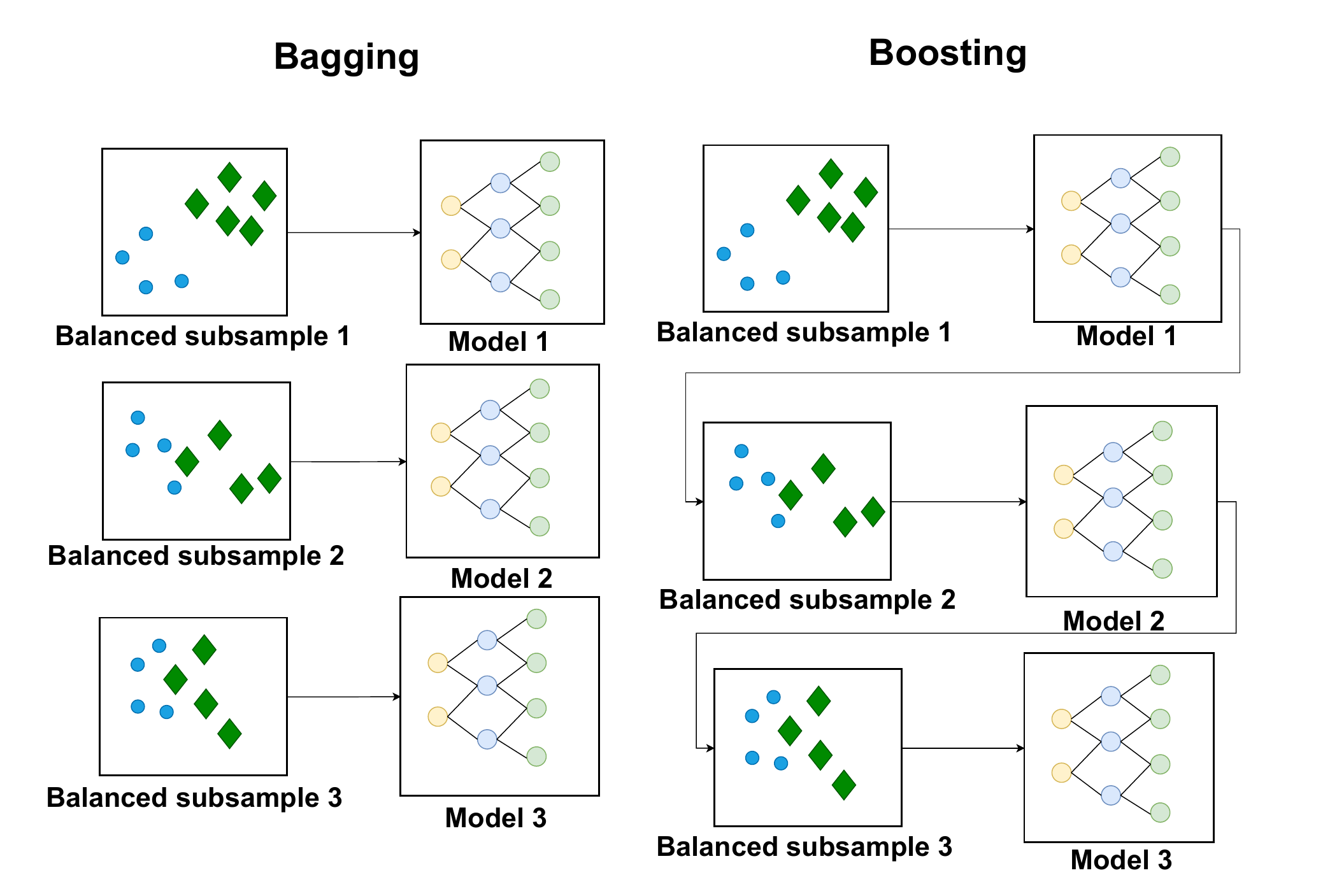}
\caption{Illustration of Bagging and Boosting. The diagram on the left shows Bagging, where the ML models are trained independently on balanced bootstrapped subsets of the data. The diagram on the right shows Boosting, where the ML models are trained sequentially.}
\label{fig:el}
\end{figure}

EL is a phenomenon in which the strengths of multiple ML algorithms, also known as weak learners, are combined to achieve a certain goal~\cite{sagi2018ensemble}. There are two primary methods through which EL can be implemented, Bagging and Boosting~\cite{liu2008exploratory}. Figure~\ref{fig:el} illustrates the fundamental concept behind them. The idea behind Bagging is that we divide our dataset into balanced bootstrapped subsets through RUS. These balanced subsets are then used to train multiple ML models, whose results are then combined through majority voting. Boosting is also similar to Bagging in the sense that the dataset is divided into balanced bootstrapped samples, but the major difference lies in the fact that the training of the models is conducted sequentially, as each ML model builds upon the errors of the previous one, thereby focusing more on difficult-to-classify instances.
EL is typically applied to tree-based algorithms, such as Decision Trees (DTs), but it can also be extended to more complex models, such as NNs. Although the core principles of EL remain unchanged, the primary difference lies in the choice of the base learner, with the tree-based models replaced by NNs.
 
\subsubsection{Balanced Random Forests (BRFs)}
Although the EL approach combines the strengths of multiple weak learners, such as DTs, this strategy can be enhanced by employing stronger learners, such as Random Forests (RF)~\cite{chen2004using}. Balanced Random Forests (BRF) are an extension of the Bagging framework, where instead of using weak learners such as DTs, we utilize the more powerful RF algorithm as learners and train them using balanced bootstrapped subsets of the dataset~\cite{chen2004using}. This modification leverages the robustness of the RF algorithm while also effectively addressing class imbalance.

\subsubsection{Balanced Epoch Training}
\begin{figure}[htbp]
\centering
\includegraphics[width=0.75\linewidth]{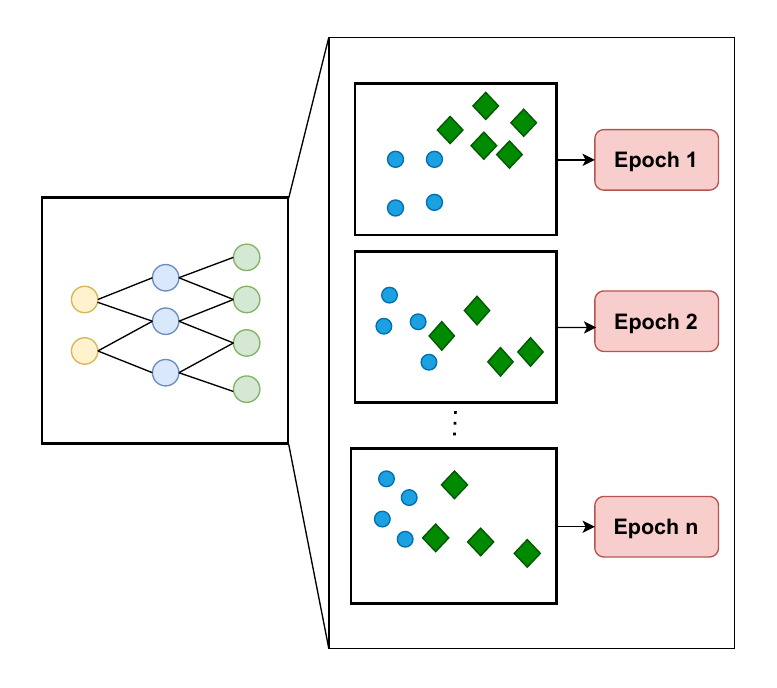}
\caption{Balanced Epoch Training where every epoch in the neural network sees a balanced subset of the dataset.}
\label{fig:bet}
\end{figure}

This technique is an adaptation of the BRF technique for use with the NN. While the EL method traditionally feeds balanced bootstrapped subsets of the dataset to each NN, this concept can be extended to the epoch level in NNs. Figure~\ref{fig:bet} shows the Balanced Epoch Training method. Specifically, this method allows us to pass a balanced bootstrapped subset of our dataset to every epoch, as seen in Figure~\ref{fig:bet}. As a result, every epoch in the NN sees a balanced distribution between classes, and thus enables even finer-grained control over class imbalance throughout the learning process.

\subsubsection{Meta-Learning}
\begin{figure}[htbp]
\centering
\includegraphics[width=0.85\linewidth]{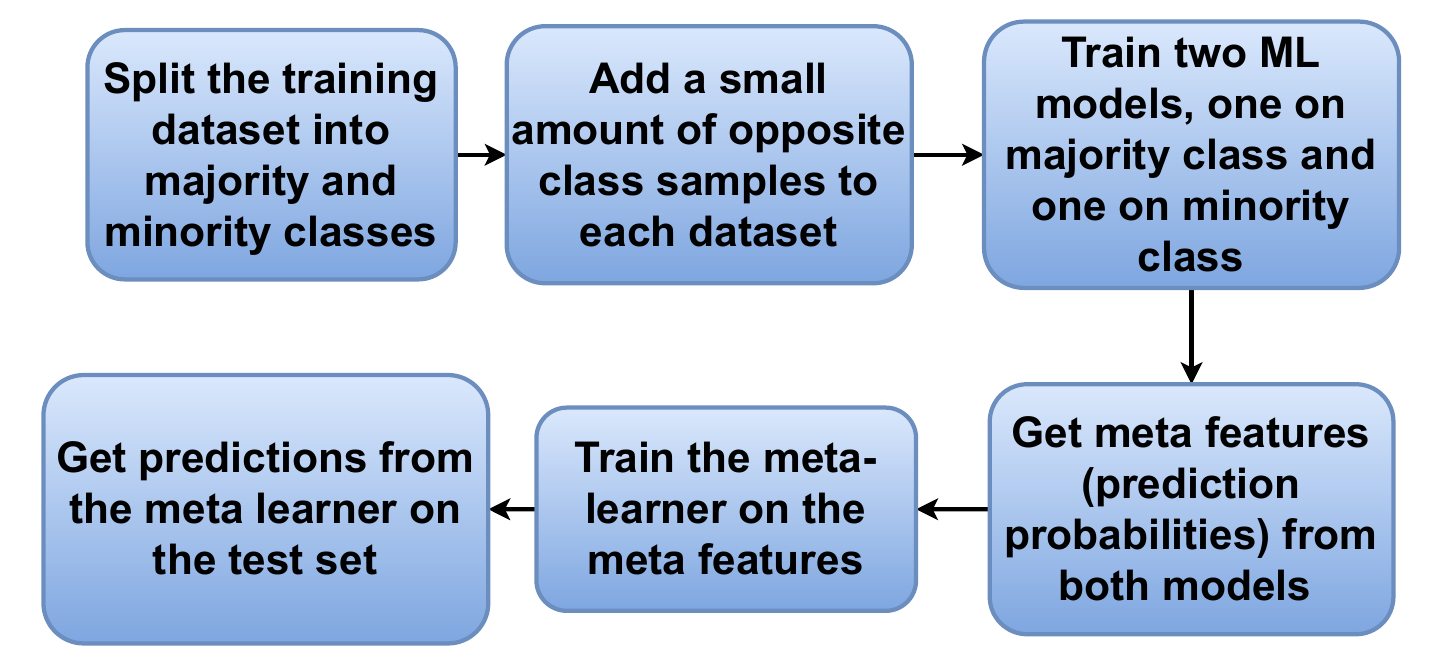}
\caption{A flowchart explaining the steps involved in Meta-Learning.}
\label{fig:meta}
\end{figure}

Meta-Learning aims to utilize insights from previous tasks to facilitate the acquisition of new and valuable information~\cite{vettoruzzo2024advances}. A similar idea can be applied to the case of class imbalance mitigation. Figure~\ref{fig:meta} shows a flow chart explaining the different steps involved in Meta-Learning. The first step is to break the dataset into majority and minority sets. A small number of samples from the opposite class are added to each dataset to avoid overfitting. Two separate ML models are then trained on the two datasets, and the predictions from these models are generated, which serve as the meta-features for the Meta-Learner. A simple ML model, such as Logistic Regression, is then used as the Meta-Learner, trained on the predictions from the previous models. The final step is to obtain the predictions from the Meta-Learner using the test set. This implementation is for a binary scenario; however, it can easily be adapted to a multi-class scenario. The only difference is that, instead of two datasets, we would need to break every class into its own dataset and then train its subsequent model. 

Model-centric methods are powerful when data is large or high-dimensional: they avoid making artificial data and often generalize better than naive over-sampling. For structured data, Weighted Learning method is a good starting point since it is the simplest to implement. BRF is a suitable technique for class imbalance in RFs, while Balanced Epoch Training (each epoch sees equal data from each class) is another practical tactic for NNs. However, there is no single approach that fits all scenarios. All the methods have to be tailored in some way or the other to accommodate the dataset.

\subsection{Post-processing}

Post-processing techniques, also known as prediction-centric approaches, are strategies that adjust the model output to favor the minority class. We identified and tested five different approaches in this category: Threshold Adjustment, Cost-sensitive Threshold, Reweighting Predictions, Probability Calibration, and Sample Weighting. The following paragraphs provide more details on these approaches. 

\subsubsection{Threshold Adjustment}
In balanced binary data sets, the prediction probability threshold is normally set at 0.5. If the prediction probability exceeds 0.5, the sample is considered to belong to the positive class; otherwise, it is considered a negative class sample. However, for imbalanced datasets, setting the threshold at 0.5 can lead to undesirable results~\cite{zou2016finding}. Threshold Adjustment is the method that is used to tune the prediction probability threshold to optimize the metric that is suitable for the target application. For the case of imbalanced datasets, the most preferred metric is the F1-score~\cite{zou2016finding}. The Algorithm~\ref{alg:threshold_adjustment} shows the pseudocode of the Threshold Adjustment approach. The idea is that we calculate the F1-score for all the different values of the threshold between 0 and 1, and select the one that maximizes the F1-score.
\begin{algorithm}[H]
\caption{Threshold Adjustment to Maximize F1-Score}
\label{alg:threshold_adjustment}
\begin{algorithmic}[1]
\Require Predicted probabilities $\hat{y}$, True labels $y$, Step size $\delta$ (e.g., 0.01)
\Ensure Optimal threshold $\tau^{*}$

\State Initialize: $\tau^{*} \leftarrow 0$, $F1_{\text{best}} \leftarrow 0$
\For{$\tau$ in $[0, 1]$ with step $\delta$}
  \State $\hat{y}_{\text{pred}} \leftarrow$
    $\begin{cases}
      1 & \text{if } \hat{y}_i \geq \tau \\
      0 & \text{otherwise}
    \end{cases} $
  \State Compute F1-score: $F1 \leftarrow \text{F1\_score}(y, \hat{y}_{\text{pred}})$
  \If{$F1 > F1_{\text{best}}$}
    \State $F1_{\text{best}} \leftarrow F1$
    \State $\tau^{*} \leftarrow \tau$
  \EndIf
\EndFor
\Return $\tau^{*}$

\end{algorithmic}
\end{algorithm}
The Threshold Adjustment approach can be adapted for the multi-class scenario by decomposing the problem into multiple one-vs-rest binary classification tasks. We treat the class for which we want to optimize the threshold as positive, and all other classes are treated as the negative class. The threshold for predicting that particular class is then optimized to maximize the selected metric. This process is repeated independently for all classes.

\subsubsection{Cost-sensitive Threshold}
In essence, the Cost-sensitve Threshold is a form of Threshold Adjustment. However, unlike traditional threshold tuning that aims to optimize evaluation metrics such as F1-score or accuracy, the goal of Cost-sensitive Threshold is to minimize the overall misclassification cost. Depending on the application, different costs are assigned to false positives and negatives according to the application context. In the failure management scenario, considering \textit{normal} as the negative class and \textit{failure} as the positive class, false negatives receive a higher cost because a failure not detected can prove critical in a real-world scenario. The threshold $\tau^{*}$ is calculated from the false positive, $C_{\text{FP}}$, and false negative, $C_{\text{FN}}$, costs according to the expression \cite{araf2024cost}:
\begin{equation}
\tau^{*} = \frac{C_{\text{FP}}}{C_{\text{FP}} + C_{\text{FN}}}.
\label{eq:cst}
\end{equation}
This threshold is subsequently applied to the prediction probabilities to achieve the optimal F1-score. 
Similar to the case of Threshold Adjustment, the Cost-sensitive Threshold method can also be extended to the multi-class case. We can treat the problem as a one-vs-rest binary classification scenario for every class and assign false-positive and false-negative costs for each class. The optimal threshold for every class is then calculated from Eq.~(\ref{eq:cst}).

\subsubsection{Reweighting Predictions}
The fundamental idea of Reweighting Predictions is to apply the class weights derived from Eq.~(\ref{eq:cw}) to the ML model's predictions, rather than incorporating them into the learning phase. The idea is to multiply the predictions from the ML model with the class weights calculated earlier, with the predictions for samples from each class multiplied by its corresponding weight. The probabilities for each sample are then normalized so that their sum equals 1. It can also happen that multiplying the predictions with weights for the classes might overestimate the probabilities. To mitigate this, a scaling factor is introduced that can be tuned to maximize the optimization metric. The scaling factor is multiplied by the weights of the classes to scale them up or down based on the context. This method is exactly applicable to the case of a multi-class scenario as well and does not require any modifications.

\subsubsection{Probability Calibration}
\begin{figure}[htbp]
\centering
\includegraphics[width=0.7\linewidth]{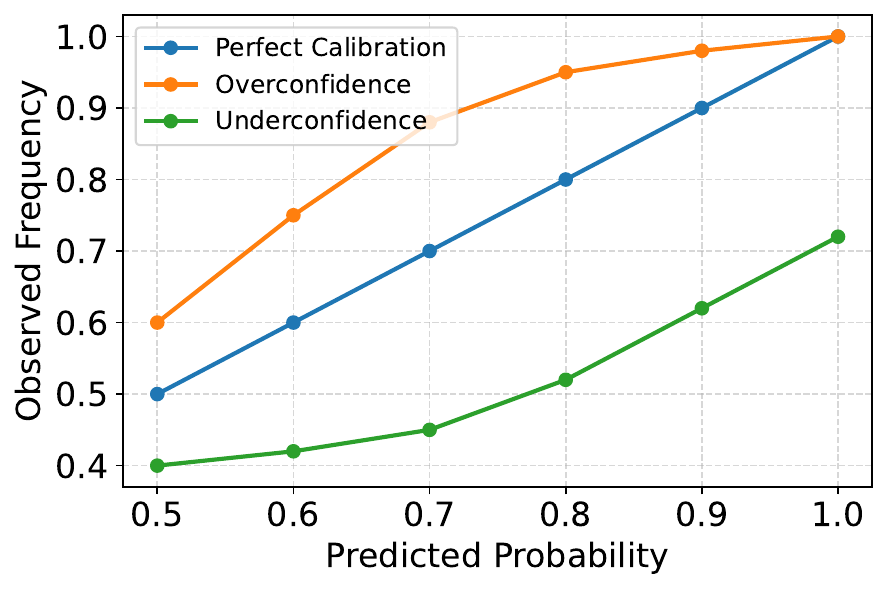}
\caption{Probability Calibration plot showing the calibration curves of perfectly calibrated, overconfident, and underconfident ML models.}
\label{fig:pc}
\end{figure}

The probabilities of an ML model are said to be well calibrated if, for any given probability score, the proportion of samples belonging to the predicted class matches that probability. For example, if the model predicts a class with 50\% confidence, then approximately 50\% of the samples in the dataset should belong to that particular class~\cite{lichtenstein1977calibration}. In some cases, it can happen that the probabilities of an ML model are not well calibrated. Figure~\ref{fig:pc} shows the calibration curves for the cases of perfect, overconfident, and underconfident calibration. In the overconfident case, the probability is higher than the actual number of samples of that class, and in the underconfident scenario, the probability is lower than the number of samples belonging to that class. Probabilities not being well calibrated can be a problem in the case of imbalanced datasets, and thus, it is essential to rectify this issue. 

There are primarily two methods for calibrating probabilities, Platt Scaling~\cite{platt1999probabilistic} and Isotonic Regression~\cite{zadrozny2002transforming}. Platt Scaling is a simple technique to tune the predicted probabilities by passing them through a sigmoid function~\cite{niculescu2005predicting}. Although the authors in~\cite{platt1999probabilistic} had introduced this method for the Support Vector Machine (SVM) algorithm, the authors in~\cite{niculescu2005predicting} have shown that it can also be applied to other ML algorithms. The disadvantage of Platt Scaling is that the transformation through a sigmoid might be beneficial to some ML models but can degrade performance for others. Isotonic Regression is a more general method where the only restriction is that the mapping function for the probabilities has to be isotonic (i.e., monotonically increasing). For the case of failure detection and identification, we employed Isotonic Regression due to its more generalized applicability. 

\subsubsection{Sample Weighting}
Sample weighting is an effective post-processing method in which we take the predictions from the ML model and identify which predictions have been misclassified. Algorithm~\ref{alg:sw} shows the pseudocode for the Sample Weighting technique. After training the initial model, the predictions from the training set are computed. Samples that have been misclassified are identified and assigned higher weights. The model is then re-trained with the updated weights and final predictions on the test set generated from the updated model.
\begin{algorithm}[H]
\caption{Sample Weighting.}
\label{alg:sw}
\begin{algorithmic}[1]
\Require Training data $(X_{\text{train}}, y_{\text{train}})$, test data $X_{\text{test}}$, model $\mathcal{M}$
\State Train initial model $\mathcal{M}_0$ on $(X_{\text{train}}, y_{\text{train}})$
\State Predict on training set: $\hat{y}_{\text{train}} \gets \mathcal{M}_0(X_{\text{train}})$
\State Identify misclassified samples: $m_j \gets \mathbb{I}(y_{\text{train},j} \neq \hat{y}_{\text{train},j})$
\State Initialize sample weights: $w_j \gets 1$ for all $j$
\State Adjust weights for misclassified samples: $w_j \gets w_j \times 2$ if $m_j = 1$
\State Retrain final model $\mathcal{M}_{\text{weighted}}$ on $(X_{\text{train}}, y_{\text{train}})$ using weights $w$
\State Predict on test set: $\hat{y}_{\text{test}} \gets \mathcal{M}_{\text{weighted}}(X_{\text{test}})$
\end{algorithmic}
\end{algorithm}
Post-processing approaches are especially useful in those scenarios where model training is fixed (e.g., a deployed model) but business priorities shift. It applies equally to structured or unstructured models. An important point to note is that these methods do not change the learned decision function; they only alter the way predictions are interpreted.

\section{Experimental Dataset and Baseline Results}
\label{sec:exper_dataset_baseline}

\subsection{Failure Detection}
\begin{figure}[htbp]
\centering
\includegraphics[width=1.0\linewidth]{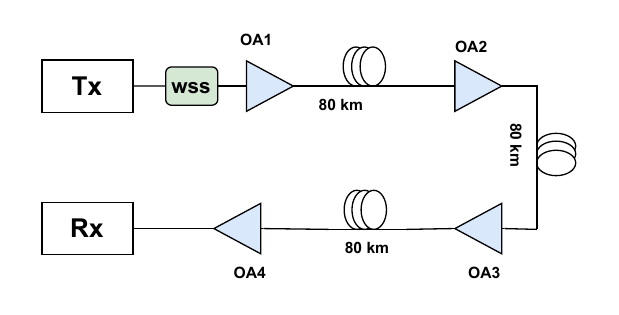}
\caption{Experimental testbed setup for failure detection.}
\label{fig:testbed}
\end{figure}

For the case of failure detection, we used the experimental datasets prepared by the authors in~\cite{silva2022learning}. These datasets were generated in the labs at the Scuola Superiore Sant'Anna University in Pisa, Italy. The experimental testbed setup is illustrated in Figure~\ref{fig:testbed}. The testbed comprises of a Transmitter (Tx) and a Receiver (Rx). A Wavelength Selective Switch (WSS) is used to emulate failures in the network, which is then followed by three Optical Amplifiers (OAs) with a total of three fiber spans spanning 80~km in length. The authors in~\cite{silva2022learning} generated two datasets from this setup. The first dataset is based on hard failures that can instantly impact the network. The second dataset is based on soft failures that gradually affect the network over time. The features collected for both datasets include the Timestamp, Type of Device, ID of the device, Bit Error Rate (BER) and the Optical Signal-to-Noise Ratio (OSNR) of Tx and Rx. For the sake of simplicity, we are considering an end-to-end monitoring scenario in which we measure the BER and OSNR of the Tx and Rx. We used the hard failure dataset to test the class imbalance mitigation techniques and then validated our results on the soft failure dataset for the case of failure detection. From now on, the hard failure dataset will be referred to as dataset 1, while the soft failure dataset will be referred to as dataset 2. 

\begin{figure*}[t]
\centering
\begin{minipage}{0.48\linewidth}
  \centering
  \includegraphics[width=0.8\linewidth]{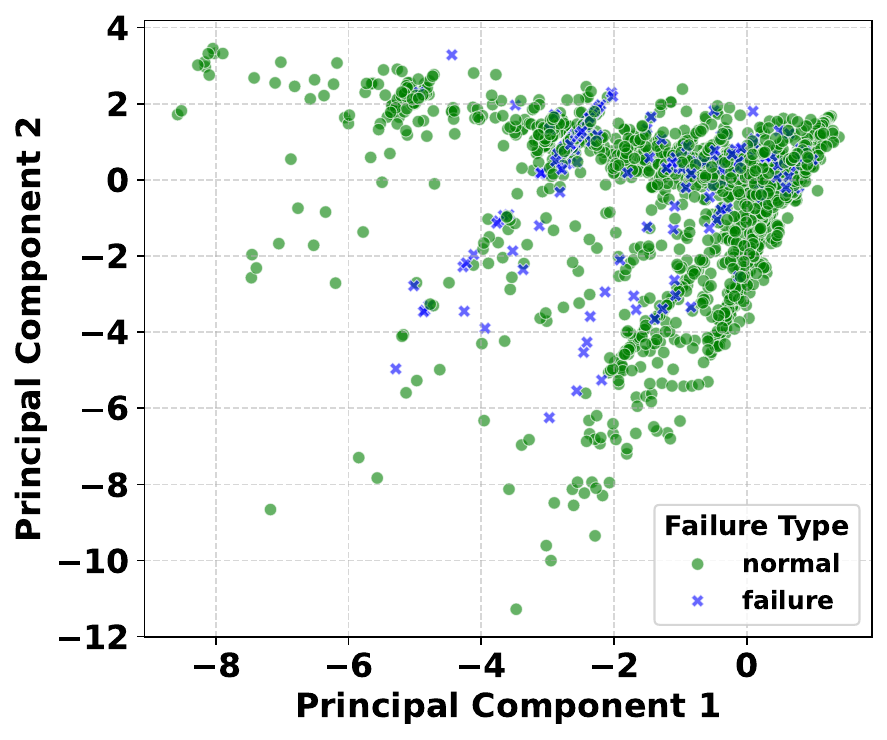}
  \caption*{(a) PCA plot for dataset 1}
\end{minipage}
\hfill
\begin{minipage}{0.48\linewidth}
  \centering
  \includegraphics[width=0.8\linewidth]{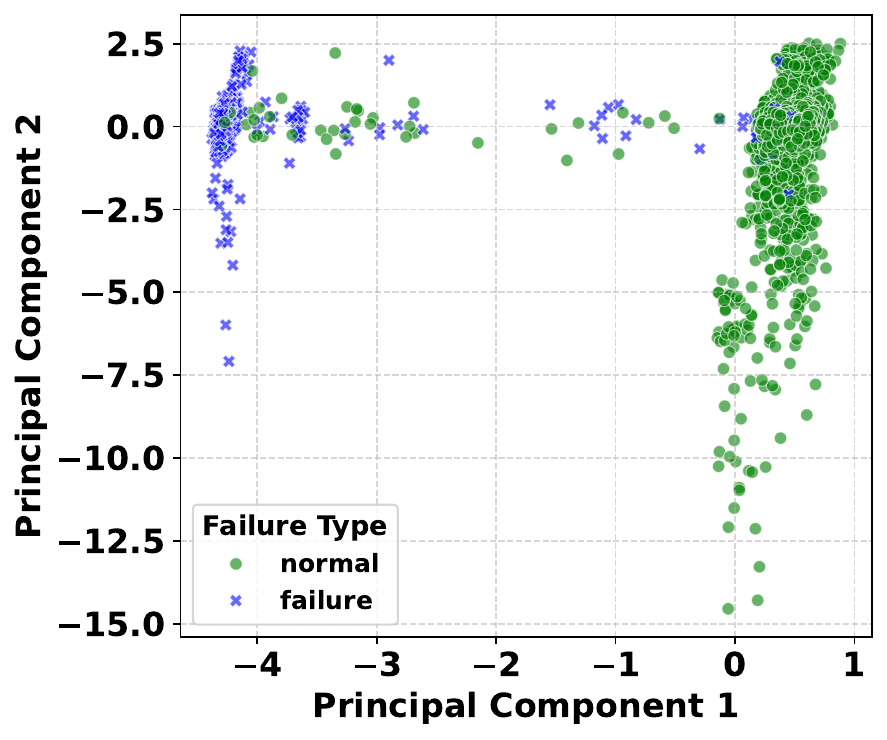}
  \caption*{(b) PCA plot for dataset 2}
\end{minipage}
\vspace{-1mm}
\caption{Principal Component Analysis for dataset 1 and dataset 2.}
\label{fig:pca}
\end{figure*}

For dataset 1, initially, there were 63248 normal samples and 2485 failure samples. After pre-processing steps and removing NaN values, we had 7859 normal samples and 194 failure samples. Similarly, for dataset 2, initially there were 9961 normal samples and 1059 failure samples, which were reduced to 6253 and 714 normal and failure samples, respectively. Figure~\ref{fig:pca} shows the Principal Component Analysis (PCA) plots for dataset 1 and dataset 2. PCA is a method of reducing the dataset from an n-dimensional dataset into a much simpler 2 or 3 dimensional space, thereby reducing the dimensionality of the dataset and preserving as much variation as possible~\cite{jolliffe2016principal}. Reducing the dimensions of the dataset helps visualize the data distribution. The feature set for dataset 1 and dataset 2 was also reduced to two components, and the PCA plot was generated. It can be clearly seen that the \textit{normal} and \textit{failure} classes have a significant amount of overlap between the classes for dataset 1, while dataset 2 has a very minimal amount of overlap. 

We can use the Fisher Discriminant Ratio (FDR) to measure the overlap between classes in a dataset. Although FDR is used primarily to assess the discriminative capacity of individual characteristics in a dataset~\cite{kalinkov2019adaptive}, we can take the average of the FDR values of all characteristics to derive a single interpretable metric that gives an idea of the general separability of the class in the feature space. A higher FDR value suggests greater separation between classes, and a lower value indicates a higher amount of overlap. FDR for a single feature can be calculated using the following formula: 
\begin{equation}
\label{eq:fdr}
\text{FDR} = \frac{\sum_{k=1}^{C} n_k \, (\mu_k - \mu)^2}{\sum_{k=1}^{C} n_k \, \sigma_k^2} ,
\end{equation}
where $C$ is the number of classes, $n_k$ is the number of samples in class $k$, $\mu_k$ is the mean class $k$ for the given feature, $\mu$ is the overall mean of the feature, and $\sigma_k^2$ is the variance of class for the given feature. The FDR value for every feature is calculated, and the overall mean is calculated to get a measure of the overlap in our dataset. The FDR values for dataset 1 and dataset 2 are 0.769 and 2.254, respectively. The lower FDR value for dataset 1 clearly shows that there is a significant amount of overlap between classes, as is also seen in Figure~\ref{fig:pca}. The higher degree of separation between classes in dataset 2 is confirmed by its higher FDR value and by Figure~\ref{fig:pca}. The varying degrees of overlap contribute to the impact of the class imbalance mitigation techniques, which will be shown in Section~\ref{sec:results}.

To establish a baseline for comparing class imbalance mitigation techniques, we selected the RF algorithm~\cite{chen2004using}. RF was chosen because of its robustness in handling imbalanced datasets, its relatively low computational complexity, and its suitability for handling simpler tasks such as failure detection, in contrast to the more complex task of failure identification. We used the scikit-learn library to implement the RF algorithm~\cite{pedregosa2011scikit}. All the RF parameters were kept the same as the default settings in scikit-learn. The baseline results obtained using the original (imbalanced) dataset are presented in Table~\ref{tab:metrics}.

\begin{table}[htbp]
\centering
\caption{\bf F1 Score for dataset 1 and dataset 2 baseline.}
\begin{tabular}{cc}
\hline
\textbf{Dataset} & \textbf{F1 Score (average)} \\
\hline
1 & 0.7659 \\
2 & 0.9470 \\
\hline
\end{tabular}
 \label{tab:metrics}
\end{table}

To evaluate performance, we employ the F1 score as the primary metric. Unlike accuracy, which can be misleading in the case of class imbalance, the F1 score provides a more balanced metric by considering both false positives and false negatives. The values in Table~\ref{tab:metrics} give the average value over 100 independent runs, thus taking into account the stochastic nature of the training process and ensuring accurate estimation of the performance variance. As is clear from Table~\ref{tab:metrics}, the baseline F1 score is relatively low for dataset 1, indicating a poor generalization likely caused by the class imbalance. The F1 score is higher for dataset 2 due to its higher degree of separation between classes, as was evident by the high FDR value, although there is still room for improvement. All subsequent experiments using mitigation techniques are evaluated relative to this baseline.

\subsection{Failure Identification}
\begin{figure}[htbp]
\centering
\includegraphics[width=1.0\linewidth]{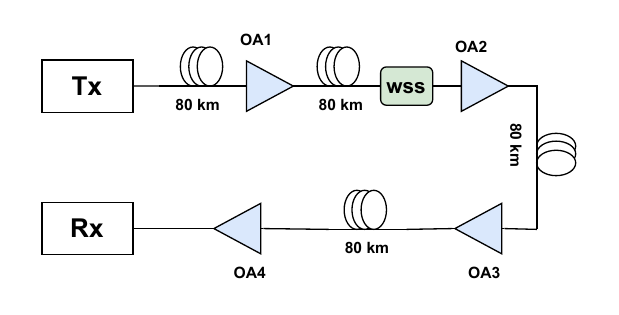}
\caption{Experimental testbed setup for failure identification.}
\label{fig:testbed_d2}
\end{figure}

In the domain of failure identification, we used the dataset employed in~\cite{khan2022data}. The experimental testbed used in this setup is shown in Figure~\ref{fig:testbed_d2}. The setup is similar to the one used for failure detection, the only difference being the number of fiber spans and the placement of the WSS. The features collected for this dataset include the input and output powers of the four OAs, the Chromatic Dispersion (CD), the Q-Factor, OSNR, and BER. Similar to the failure detection case, we are considering an end-to-end system where we only monitor the BER and OSNR.

\begin{table}[htbp]
\centering
\caption{\bf Number of samples in different classes for dataset 3.}
\begin{tabular}{ccc}
\hline
\textbf{Failure Label} & \textbf{Number of samples} & \textbf{Failure Type}\\
\hline
0 & 2184 & Normal \\
1 & 224 & Filter Tightening \\
2 & 152 & Attenuation \\
3 & 991 & Filter Tightening \\
&&+ Attenuation \\
4 & 254 & Filter Tightening \\
&&+ Filter Shift \\
5 & 325 & Filter Shift \\
\hline
\end{tabular}
 \label{tab:data_d2}
\end{table}

\begin{figure}[htbp]
\centering
\includegraphics[width=0.8\linewidth]{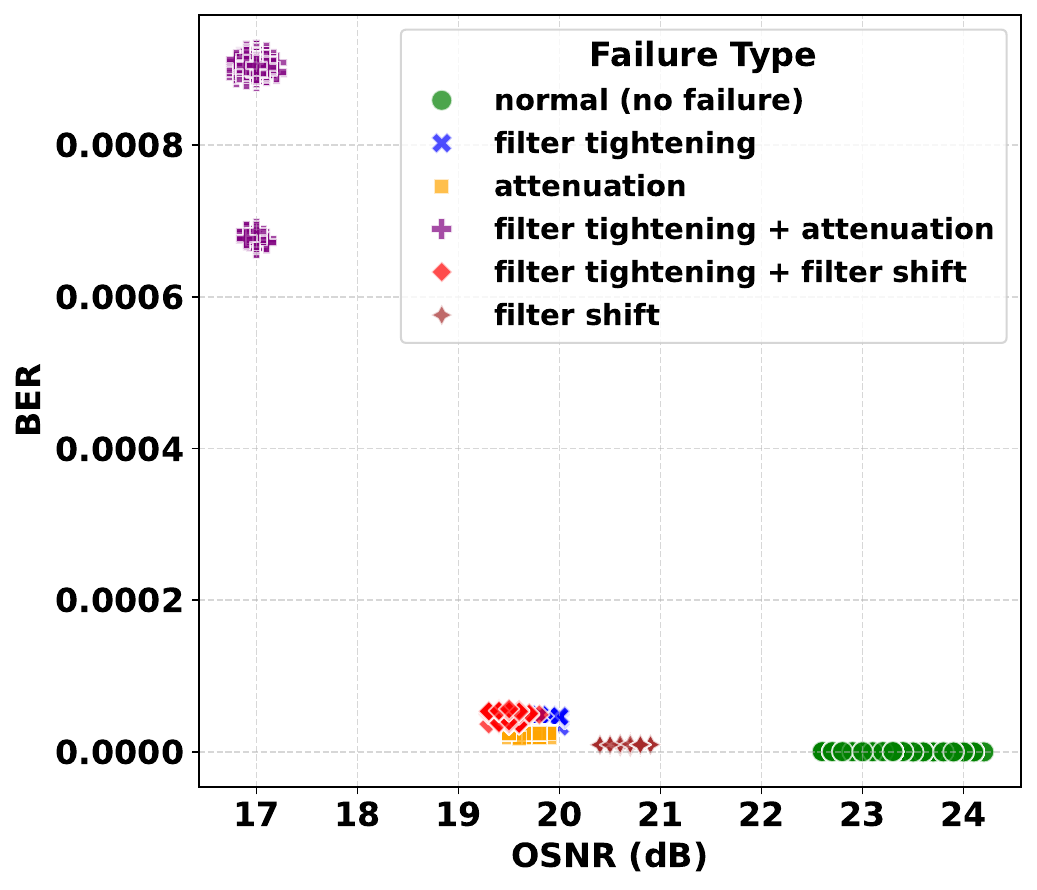}
\caption{Feature distribution by class for dataset 3.}
\label{fig:pca_d2}
\end{figure}

The total number of samples in the different classes and the description of the different failures are reported in Table~\ref{tab:data_d2}. It can be clearly seen that there is a significant amount of class imbalance between the classes, especially classes 1, 2, and 4. We will refer to this dataset as dataset 3 from this point onward. The plot of the distribution of the characteristics for data set 3 is shown in Figure~\ref{fig:pca_d2}, indicating that there is a significant overlap between failures 1, 2 and 4 while the other three classes show good separation between them. This is also confirmed by the very high FDR value of 181.2 for this dataset.
For the evaluation of dataset 3, we used the NN algorithm. The reason for using NN was that failure identification is a relatively more complex task than failure detection, and thus, NN's ability to identify more complex patterns among features comes in handy. Our baseline NN has the following parameters: an input layer of 2 neurons (corresponding to the two features), two hidden layers of 20 and 10 neurons, respectively, an output layer of 6 neurons (corresponding to the number of classes), a learning rate of 0.0001, Adam as the optimizer, and a batch size of 8. These are the same parameters that the authors in~\cite{khan2022data} used in their paper. The original dataset in the baseline NN yielded an F1 score of 0.7751 (average score over 100 independent runs), which is on the lower side and shows that, due to the class imbalance in our dataset, the NN is not able to perform optimally. The class imbalance mitigation techniques for the failure identification case will be compared with this baseline in Section~\ref{sec:results}.

\section{Results}
\label{sec:results}

\subsection{Failure Detection}
\begin{figure*}
\centering
\includegraphics[width=0.9\linewidth]{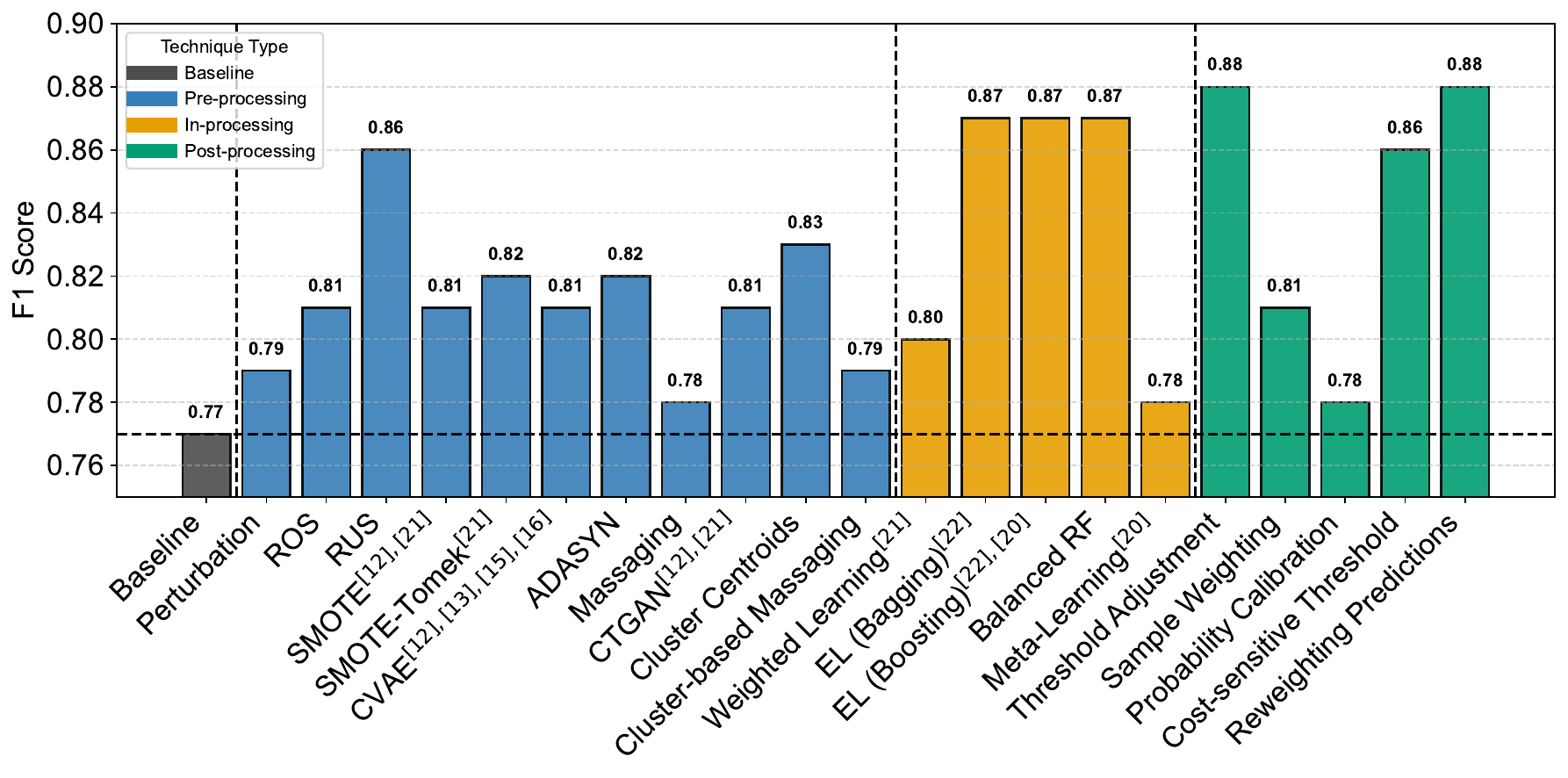}
\caption{F1 score comparison of class imbalance mitigation techniques for dataset 1. Values in the brackets indicate the scientific papers that have used these techniques in the context of failure detection/identification in optical networks.}
\label{fig:complete_results}
\end{figure*}

Figure~\ref{fig:complete_results} shows the average F1 scores for Data Set 1 after applying the different class imbalance mitigation techniques mentioned in Section~\ref{sec:class_imb_miti}. In all categories, the methods performed consistently above the baseline established in the previous section. The results of the F1 score are shown here due to their suitability for imbalanced classification tasks, but metrics such as precision, accuracy, and recall also reported improvements.

Among preprocessing methods, we observe that RUS techniques yielded the most improvement, with a 12\% increase in the F1 score compared to the baseline. The heuristic-based label flipping methods showed minimal improvements, primarily due to the fact that flipping the labels might introduce more noise in the dataset and alter the data distribution, which can ultimately lead to suboptimal generalization. The over-sampling approaches showed almost the same level of improvement, while the GenAI methods also struggled to make much impact. The inability of GenAI methods to significantly improve the F1 score is explained by the low FDR score (0.769) of this dataset. There is a significant amount of overlap between classes in the dataset, and thus GenAI methods struggle to generate good-quality samples.

Across the in-processing category, the EL techniques and the BRF give the most optimal result, providing an improvement of up to 13. 6\% compared to the baseline. Weighted Learning and Meta-Learning provided only minimal improvement over the baseline. The EL and BRF methods outperformed RUS, the best-performing approach in the pre-processing category. Moving towards the post-processing domain, Threshold Adjustment and Reweighting Predictions provide the most notable improvement, increasing the F1 score by up to 15.3\% over the baseline. The Cost-sensitive Threshold method also gives impressive performance, while Sample Weighting and Probability Calibration were only able to offer minimal improvement. This analysis provides significant insight into the advantages offered by post-processing approaches in the failure detection area. Threshold Adjustment was the best-performing method among all categories, and its main advantage lies in the fact that it operates on the predictions from the ML model, thereby reducing dependence on the data quality.

\begin{figure}[htbp]
\centering
\includegraphics[width=1.0\linewidth]{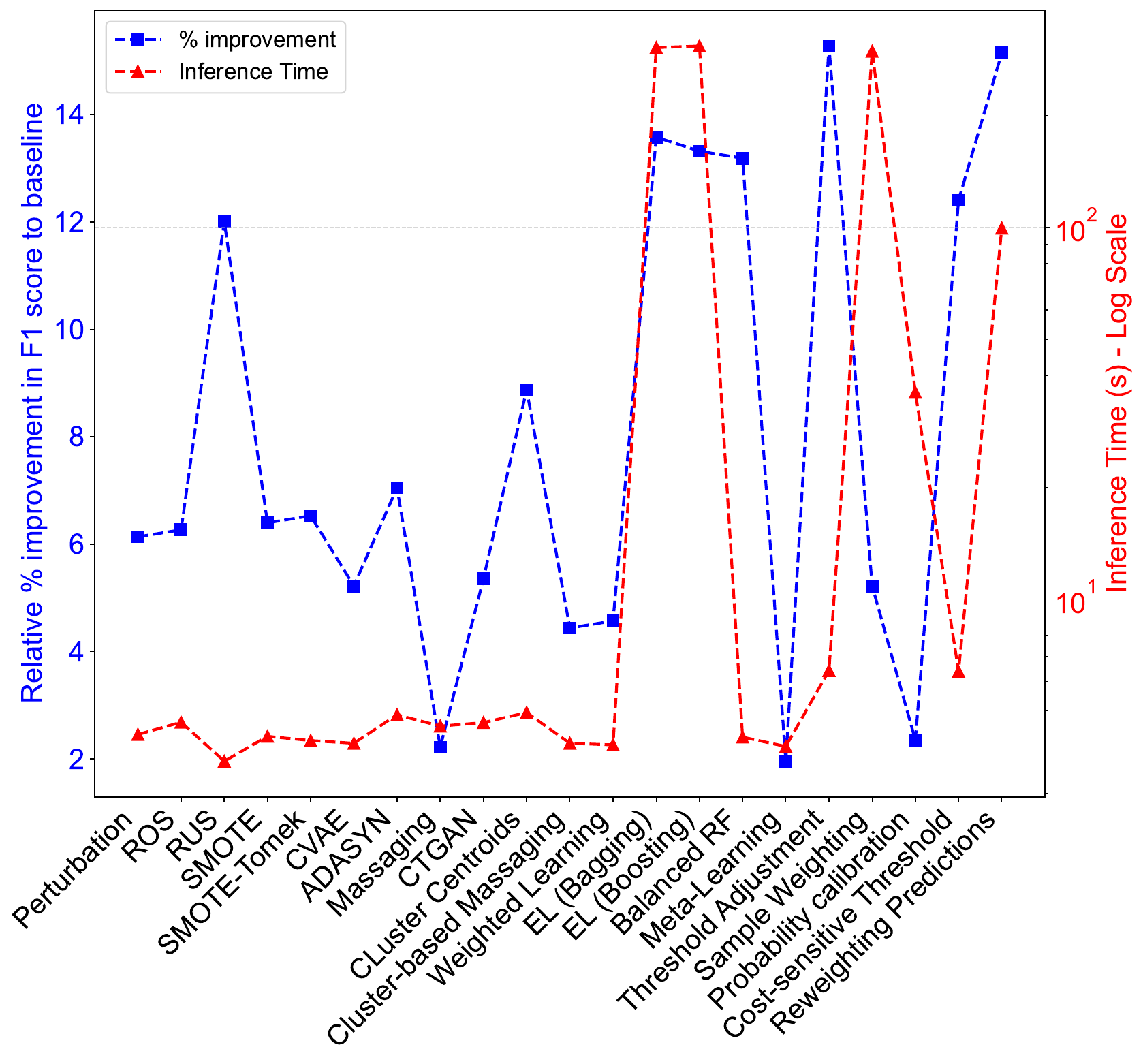}
\caption{A dual-axis plot showing the percentage improvement for the class imbalance mitigation techniques compared to the baseline (blue axis) and the inference times for the techniques (red axis).}
\label{fig:percent_improv_run_time}
\end{figure}

Figure~\ref{fig:percent_improv_run_time} shows a dual-axis plot depicting the relative percentage improvement in the F1 score for the class imbalance mitigation techniques (blue y axis) and the inference times (red y axis). We have chosen to show the inference times here rather than the training times because inference times are much more critical in an optical network setting. The most notable observation in Figure~\ref{fig:percent_improv_run_time} is that the inference times for the pre-processing and in-processing approaches are comparable to the baseline, and this is because these methods work at the data or algorithm level and do not introduce additional computational steps during the inference phase. Figure~\ref{fig:percent_improv_run_time} shows the trade-off between the percentage improvement and the inference time in all the class imbalance mitigation techniques. Focusing on the left-hand side, among the pre-processing methods, the RUS technique has the lowest inference time, mainly due to the reduced size of the dataset, which then results in a much less complex model. Moving towards the middle, the in-processing categories show an upward trend in the inference time. In particular, EL approaches incur a significant rise in inference time, and the reason for this is that, since during training time multiple ML models are being combined in an ensemble fashion, the complexity of the model increases, and thus necessitates additional time for inference. Continuing to the post-processing methods, we observe an increase in the inference times across all the techniques compared to the pre-processing approaches. This was expected because all the computation was performed during the inference phase for the post-processing techniques. 

The key takeaway from this result is that there is a trade-off between higher performance and increased computational overhead. As we move from left to right in the Figure (from pre-processing to post-processing), the performance improvement increases for some methods, but we also observe rising inference times for the same techniques. Therefore, in the context of failure detection, the method to choose is based on the preference between computational time and performance improvement. In latency-critical applications, techniques such as RUS might be more beneficial due to the lesser inference time, but in areas where performance is more critical, Threshold Adjustment might be the preferred approach.

\begin{figure}[htbp]
\centering
\includegraphics[width=0.8\linewidth]{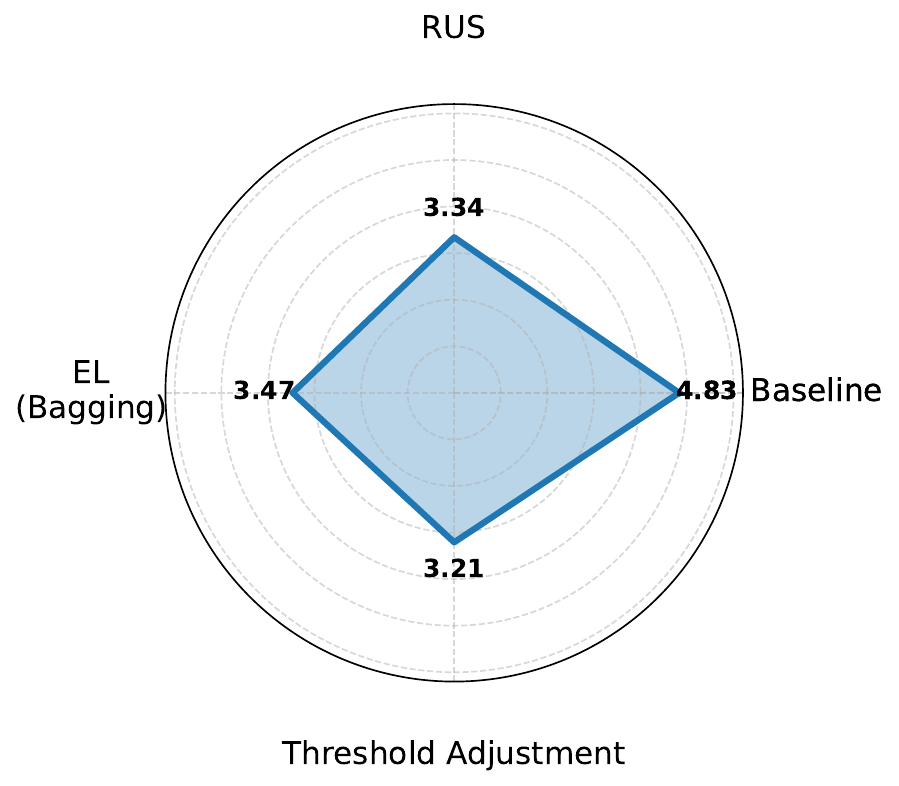}
\caption{Variance-to-mean ratio (VMR) of the best-performing method in all class imbalance mitigation categories.}
\label{fig:variance_plot}
\end{figure}

An essential step in our approach was to examine the variance in the results due to the stochastic nature of the ML algorithms. To this end, Figure~\ref{fig:variance_plot} shows the variance-to-mean ratio (VMR) for the best performing techniques in each category for 100 individual runs. It can be seen that the VMR value for RUS, Bagging, and Threshold Adjustment is lower than that of the baseline, and Threshold Adjustment has the lowest overall VMR. This is a key finding because it demonstrates that class imbalance mitigation techniques provide stable and consistent results in addition to improvement in average F1 scores. 

\textbf{}

\begin{figure*}
\centering
\includegraphics[width=1.0\linewidth]{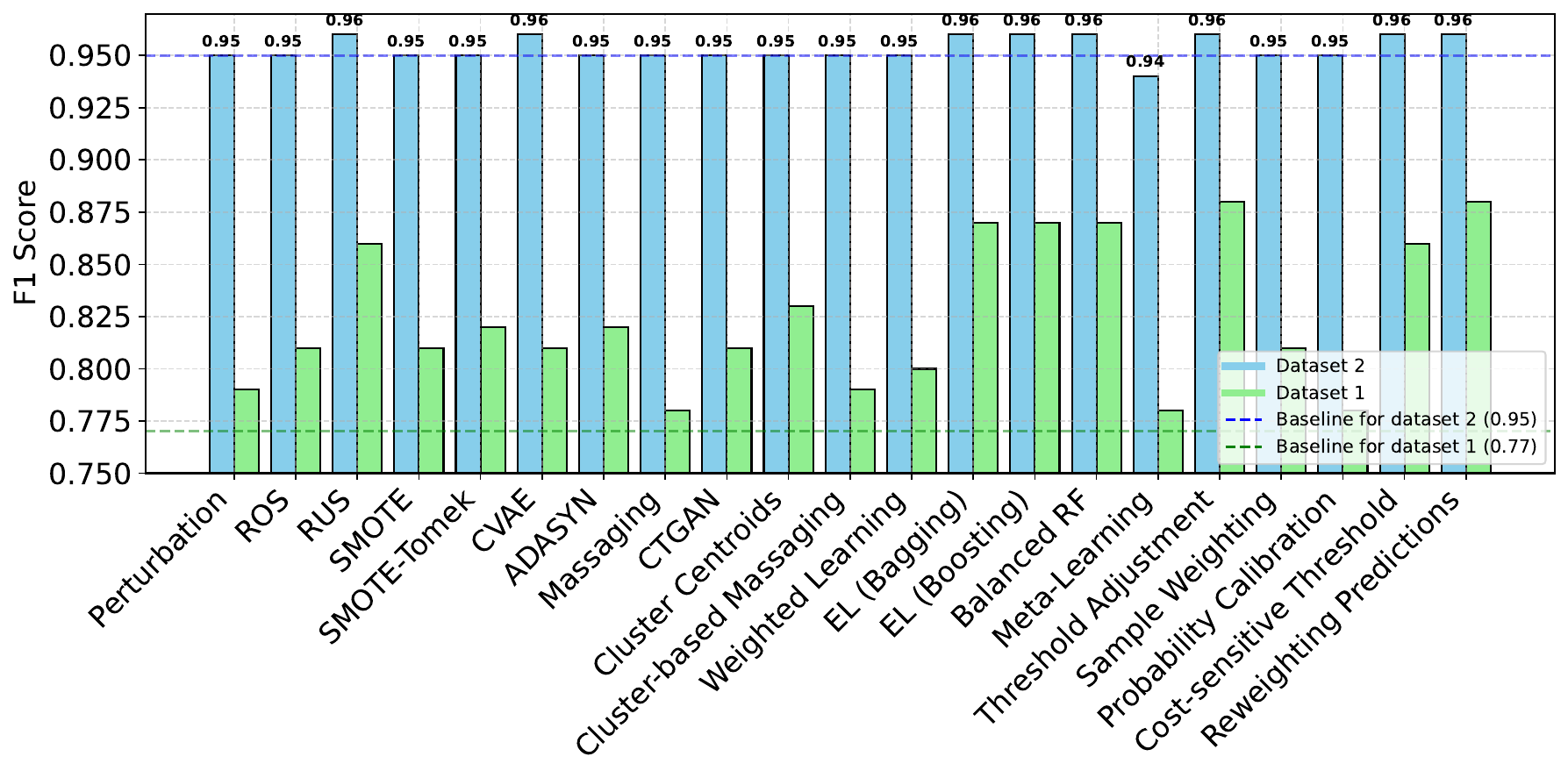}
\caption{Comparison of F1 scores for the class imbalance mitigation techniques for datasets 1 and 2. The values at the top of the blue bar are the F1 scores for the class imbalance mitigation techniques for dataset 2. The F1 scores for dataset 1 are not listed at the top of the bars in this Figure as they have already been listed in Figure~\ref{fig:complete_results}.}
\label{fig:complete_results_comp}
\end{figure*}

The next part of our analysis was to validate the results obtained in Figure~\ref{fig:complete_results} on dataset 2. Figure~\ref{fig:complete_results_comp} shows the comparison of the F1 scores for dataset 1 and dataset 2 for all the class imbalance mitigation methods. If we look at the results for dataset 2 (blue bars), we can see that among the pre-processing approaches, the CVAE and the RUS methods provide the most expressive improvement in terms of the F1 score. The increased effectiveness of GenAI methods for dataset 2 can be understood by its higher FDR value (2.254). A higher FDR value indicates greater separation between classes, which was also confirmed by the PCA plot in Figure~\ref{fig:pca}. This increased separation results in the GenAI model generating good-quality synthetic samples without altering the data distribution. This validates our earlier argument that GenAI approaches rely on the quality of the dataset to provide good-quality synthetic data. The other techniques in the pre-processing domain performed on similar lines as was observed for dataset 1 and exhibited similar F1 scores.

Moving to the in-processing domain, we observe very similar trends to what we saw for the case of dataset 1. EL methods and BRF are the best-performing techniques, providing the most improvement in terms of the F1 score. Similar to dataset 1, Weighted Learning and Meta-Learning led to similar results as compared to the baseline. Threshold Adjustment, Cost-sensitive threshold, and Reweighting Predictions are the best-performing techniques in the post-processing category, as was the case for dataset 1. The performance of Sample Weighting and Probability Calibration methods was comparable to the baseline. These results validate the trends we observed in Figure~\ref{fig:complete_results}, as the Threshold Adjustment method provides the most expressive improvement in the F1 score in dataset 1 as well as dataset 2 among all the class imbalance mitigation techniques. In the in-processing domain, the EL and BRF methods provided the most improvement for both datasets. The only difference observed was in the pre-processing category, where RUS performed best for dataset 1, while CVAE and RUS were the best techniques for dataset 2. This result confirms the argument that the post-processing methods reduce the reliance on the dataset's quality, as we observed similar results for two datasets of varying separability among the classes.

Another very important insight that can be generated from the result in Figure~\ref{fig:complete_results_comp} is that for a binary classification task that is characterized by high separability between the two classes (such as dataset 2), the effort required to achieve optimal performance is substantially reduced, as was evident by the already high F1 score for the baseline of dataset 2. In such problems, once a sufficient number of samples from each class is obtained that characterizes the distribution of the two classes, the model is robust against class imbalance. Therefore, the addition of more majority class samples does not make the model biased towards the majority class as long as they do not change the data distribution and reduce the margin of separation between the two classes.

\subsection{Failure Identification}
\begin{figure*}
\centering
\includegraphics[width=1.0\linewidth]{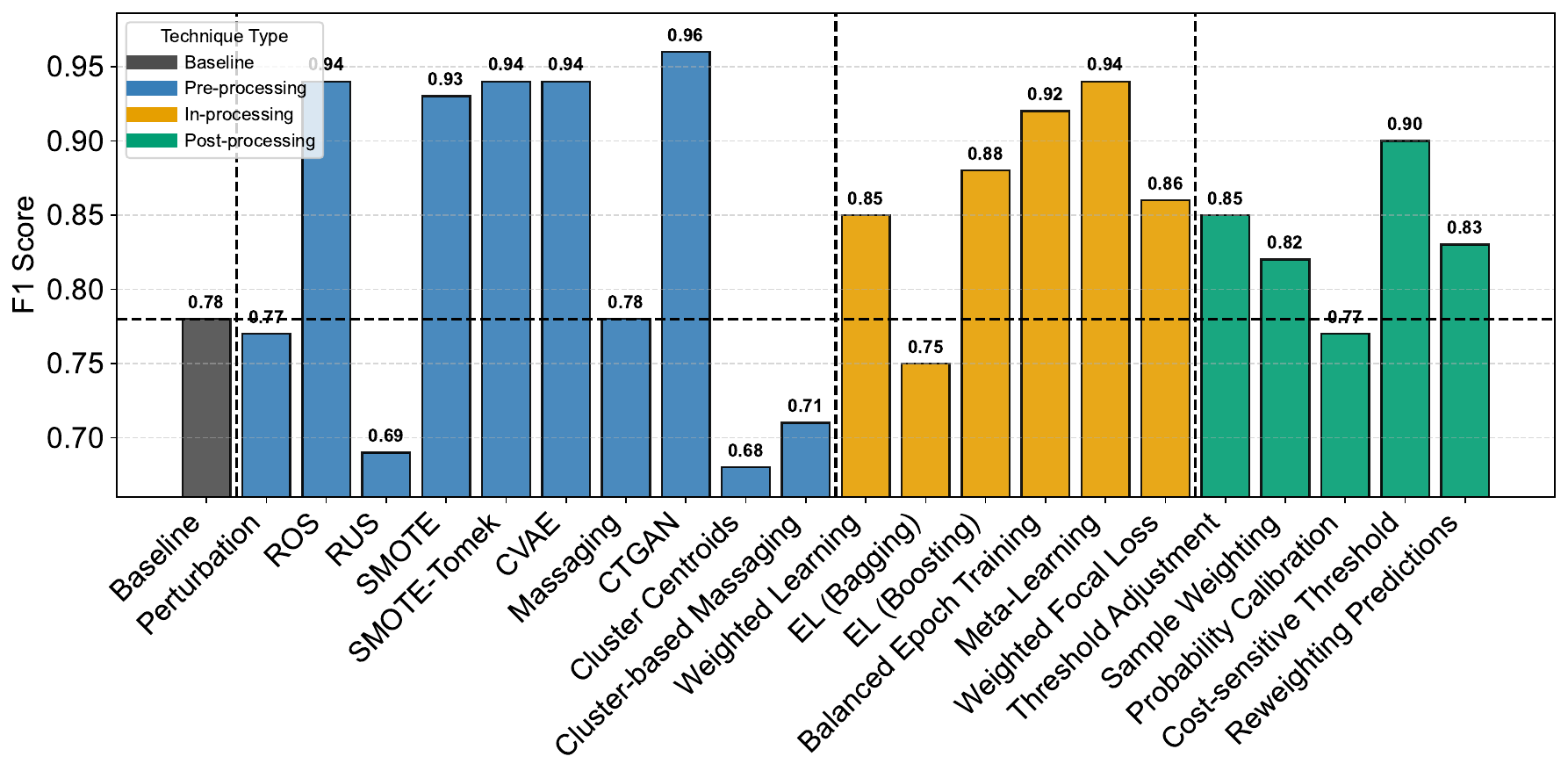}
\caption{Performance of class imbalance mitigation methods for dataset 3 for the case of failure identification.}
\label{fig:complete_results_fi}
\end{figure*}

The performance of class imbalance mitigation techniques using dataset 3 for failure identification is shown in Figure~\ref{fig:complete_results_fi} implemented using the NN architecture mentioned in Section~\ref{sec:exper_dataset_baseline}. In the pre-processing category, we can see that GenAI methods, specifically CTGAN, have performed the best, providing an improvement of up to 24.2\%. This can be attributed to the fact that the FDR value for dataset 3 is very high (181.2), indicating a higher degree of separation between classes, also confirmed by the feature distribution plot in Figure~\ref{fig:pca_d2}. The higher degree of separation between classes results in the GenAI models being trained well and generating quality synthetic samples, leading to an excellent improvement in the F1 score. The over-sampling methods, such as ROS and SMOTE, also performed well, indicating that the NN greatly benefited from more minority class samples. The heuristics-based label flipping techniques led to slight improvement or even performance degradation in the case of Cluster-based Massaging. This is because dataset 3 is highly separable, so flipping labels only introduces noise in the dataset. Deciding which labels to flip is a complicated scenario in a multi-class case as compared to a binary case. Flipping the wrong labels can lead to performance degradation. RUS performed poorly in the case of failure identification, primarily due to the reduction of data samples for NNs, as they require a good amount of data to be trained optimally. One thing to note is that the ADASYN method did not work for this dataset because ADASYN focuses on the samples at the boundary, and since dataset 3 is very separable, it could not find any boundary samples to over-sample the data.

The Meta-Learning algorithm performed best for the in-processing approaches, resulting in an improvement of 21.7\% over the baseline. The Balanced Epoch Training and Boosting methods also provided a considerable improvement in the F1 score. The Weighted Learning and WFL methods performed almost identically, with the WFL technique providing only a marginal improvement over Weighted Learning. The most interesting result was the performance degradation experienced with the Bagging technique. This can be attributed to the fact that each learner in the ensemble is trained on an under-sampled subset of the data. As a consequence, individual NNs are exposed to limited samples of the dataset, which leads to insufficient training. Since the final prediction is derived through majority vote in these suboptimal learners, the overall performance of the ensemble is compromised. 

Among post-processing techniques, the Cost-sensitive Threshold approach performed the best, increasing the F1 score by 16.4\%. Threshold Adjustment was the second-best method in the category, while Sample Weighting and Reweighting Predictions resulted in similar performance. The Probability Calibration approach was on the same level as the baseline, mainly because the SoftMax probabilities of the output layer of NN are already well calibrated, and thus applying calibration techniques results in little to no improvement. We observe here that the post-processing methods do not provide as expressive an improvement as in failure detection. This is because post-processing approaches are trickier to apply in a multi-class scenario than in a binary scenario, such as failure detection. In Threshold Adjustment, adjusting the threshold of more than one class can lead to performance degradation for other classes. Similarly, in Sample Weighting, adjusting the weights for the samples of one class can influence the performance of the other classes. Even in the best-performing method, Cost-sensitive Threshold, each class had to be assigned different costs for false positives and false negatives, and thus, very precise fine-tuning of all the costs was needed to prevent performance degradation for any of the classes.

\begin{figure}[htbp]
\centering
\includegraphics[width=1.0\linewidth]{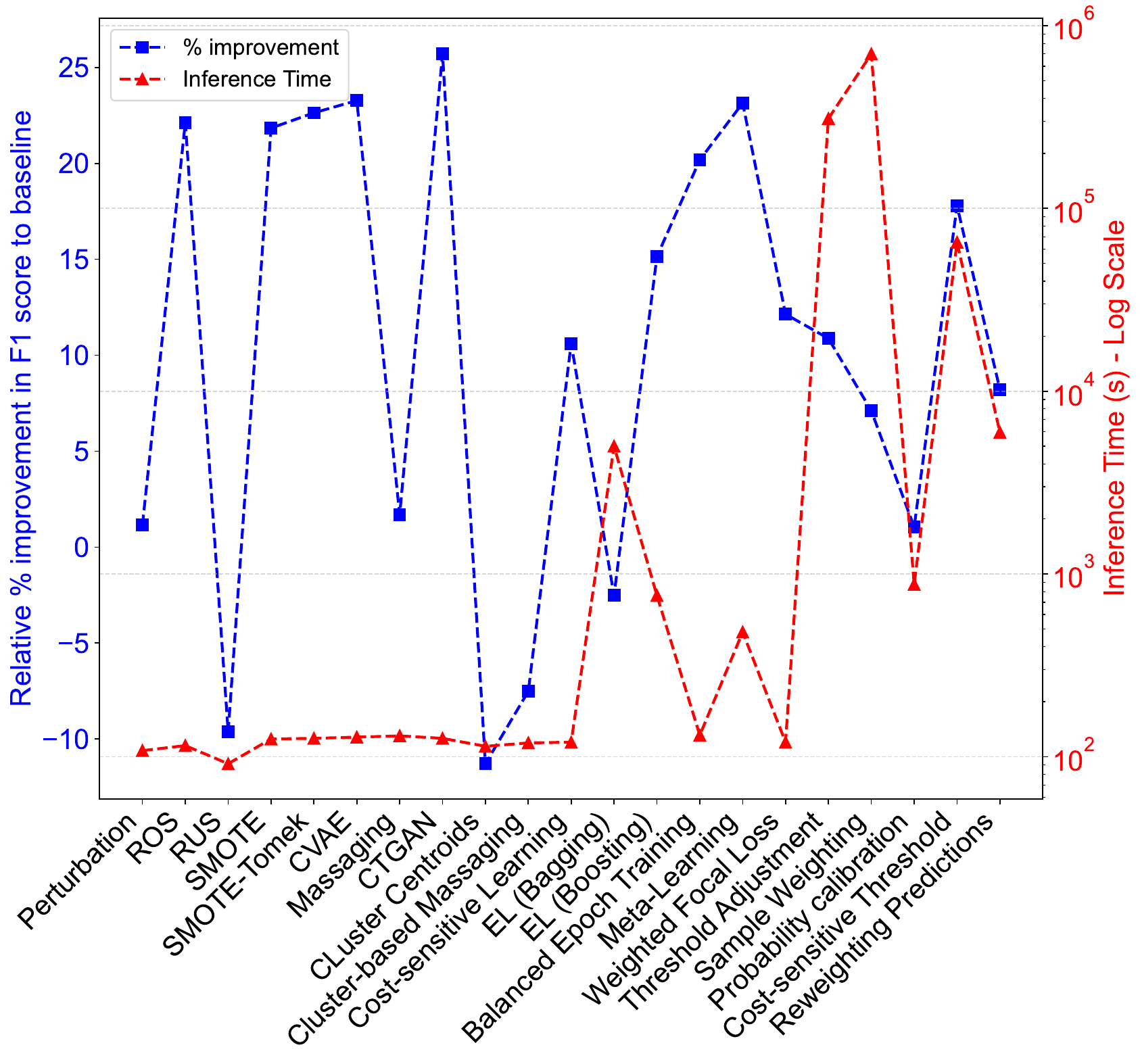}
\caption{Dual-axis plot showing the percentage improvement in F1 score relative to baseline (blue y axis) and inference times for the class imbalance mitigation techniques for dataset 3.}
\label{fig:improv_run_time_fi}
\end{figure}

Figure~\ref{fig:improv_run_time_fi} shows a dual-axis plot for the relative performance improvement (blue y axis) and the inference times (red y axis) after the application of the class imbalance mitigation techniques for dataset 3. The trend observed in this plot is similar to the one in Figure~\ref{fig:percent_improv_run_time}. As we move from left to right of the Figure (from pre-processing to post-processing techniques), most of the techniques show an increase in inference times. The post-processing methods showcase the highest inference times, whereas the pre-processing approaches have the least inference times. Moreover, all the pre-processing methods have almost similar inference times, showing that most of the computation occurs during training time. There is also a general reduction in performance improvement as we move from pre-processing to post-processing, which shows that in the failure identification case, the pre-processing techniques, more specifically GenAI methods, are ideal to use in both instances; where latency is critical and where performance is mandatory, as we can see from the results that the performance of CTGAN is the best and the inference time is also very minimal.

\begin{figure}[htbp]
\centering
\fbox{\includegraphics[width=1.0\linewidth]{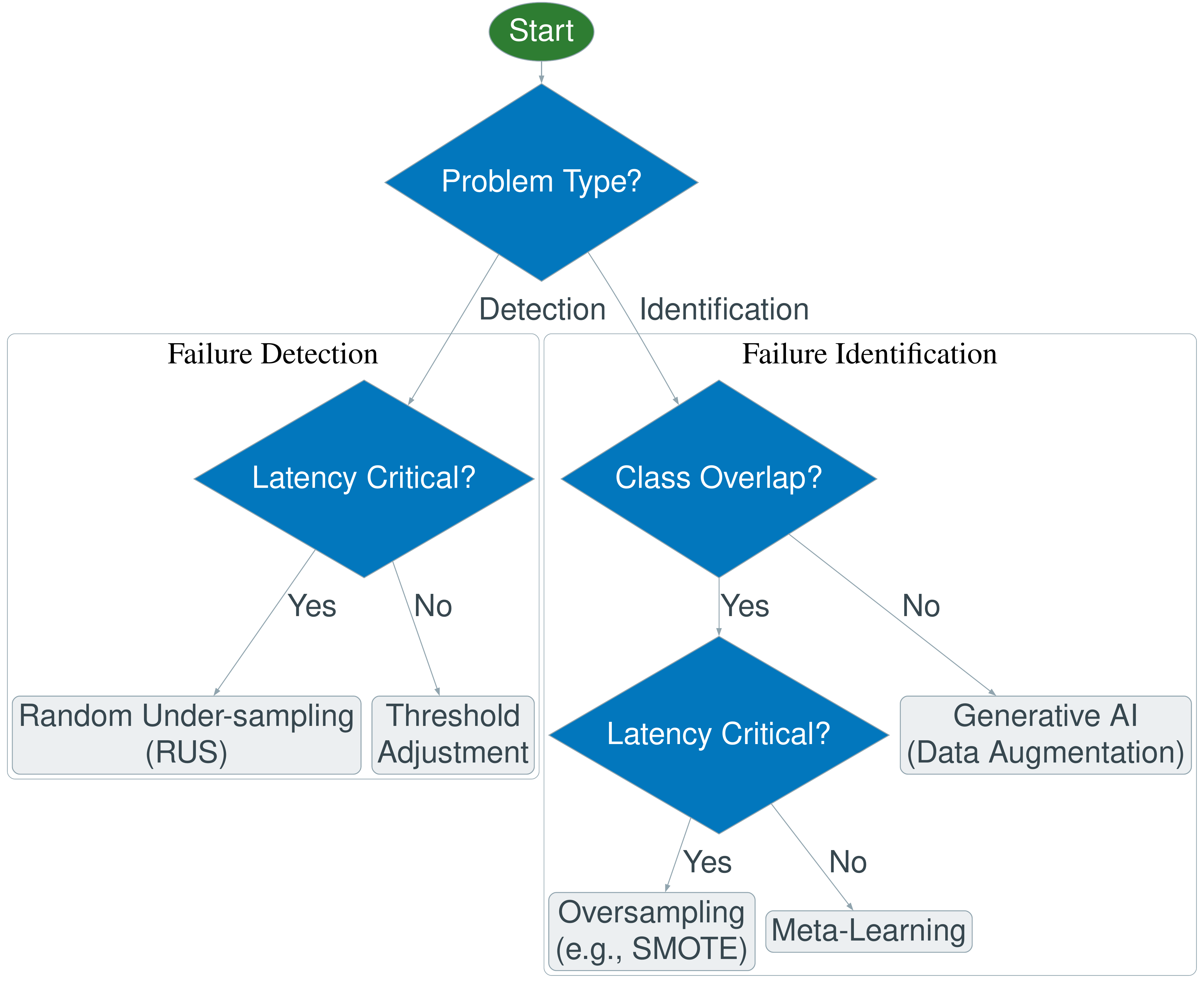}}
\caption{Flow chart outlining the blueprint to select methods to mitigate class imbalance for failure detection and identification.}
\label{fig:flow}
\end{figure}

Finally, Figure~\ref{fig:flow} shows a flow chart in which we assemble all our findings and provide a blueprint that can be used to mitigate the class imbalance problem for the detection and identification of failures. The most important aspects to consider here are whether the classes in the dataset contain any overlap and whether the task is sensitive to latency. Focusing on the failure detection task, the choices are simpler to make. Regardless of whether there is class overlap or not, if latency is critical, the RUS is the best approach to be used because of the shorter inference time and good performance gains. If computational efficiency is not as essential and maximizing performance is the ultimate goal, the Threshold Adjustment technique is the ideal choice due to the significant performance gains achieved because it relies less on the quality of the dataset and works directly on the predictions of the model. 

Considering the case of failure identification, in the event that class overlap and inference time are relevant, over-sampling methods such as SMOTE can be considered, which provided impressive performance in terms of the F1 score, in addition to keeping the inference times significantly low. If latency is not that important, Meta-Learning will be able to provide more performance improvement than over-sampling methods. In the case of no class overlap, GenAI approaches will be the best to use, regardless of whether the system is latency critical or not, as they provide the most expressive performance improvement for imbalanced datasets when the datasets are of good quality and the class overlap is minimal.

\section{Conclusion}
\label{sec:conclusion}
As applications and services that rely on optical networks become increasingly latency sensitive, the goals of network design have shifted towards the minimization of time delays in addition to maximizing throughput. This, in turn, leads to the increasing importance of fast evaluation of network failures. This paper presents the most extensive evaluation to date of pre-processing, in-processing, and post-processing techniques for class imbalance mitigation in optical network failure detection and identification, and the first systematic assessment of post-processing methods in this domain. Using experimental datasets with varying class separability, we show that Threshold Adjustment offers the highest improvement in the F1 score for failure detection (up to $+$15.3\%), with reduced dependence on the quality of the dataset, but at the cost of a longer inference time. For latency-critical scenarios, RUS offers a better balance between accuracy and speed.

Our results also reveal that the best-performing methods in each category improve not only accuracy but also stability of the model, reducing variance relative to the baseline. For failure identification, GenAI methods (CVAE, CTGAN) and over-sampling approaches achieve gains of up to 24.2\%, whereas post-processing techniques are less effective and incur significant inference delays. We recommend over-sampling (e.g., SMOTE) for latency-sensitive, high-overlap scenarios; Meta-Learning when latency is less critical; and GenAI methods for well-separated classes.

These findings provide actionable guidance for selecting imbalance mitigation strategies under varying operational constraints in optical networks. We anticipate that future work will extend the analysis to other failure management tasks, such as localization, and validate findings on real-world datasets, addressing the current data scarcity challenge in this field.

\section*{Funding}This work has received funding from the European Commission MSCA-DN NESTOR project (G.A. 101119983), which is received via UKRI (EPSRC, Grant Ref: EP/Y031024/1) in the UK. 

\section*{Acknowledgments}Sergei Turitsyn acknowledges EPSRC project TRANSNET (EP/R035342/1). João Pedro acknowledges the HORIZON \text{ALLEGRO} project (GA 700001837). \text{Mohammad} Hosseini and Antonio Napoli acknowledge the HORIZON SENSEI project (GA No. 101189545).

\bibliography{sample}

\end{document}